\documentclass[letterpaper]{article} %
\usepackage{aaai23}  %
\usepackage{times}  %
\usepackage{helvet}  %
\usepackage{courier}  %
\usepackage[hyphens]{url}  %
\usepackage{graphicx} %
\usepackage{natbib}  %
\usepackage{caption} %
\usepackage{algorithm}
\usepackage{algorithmic}
\usepackage{times}
\usepackage{epsfig}
\usepackage{graphicx}
\usepackage{xcolor}
\usepackage{amsmath}
\usepackage{amsfonts}
\usepackage{makecell}
\usepackage{algorithm}
\usepackage{algorithmic}
\usepackage{float}
\usepackage{tabularx}
\usepackage{longtable}
\usepackage{supertabular}
\usepackage{xltabular}
\usepackage{array}
\usepackage{colortbl}
\usepackage{tcolorbox}
\usepackage{adjustbox}
\usepackage{amssymb}
\usepackage{soul}
\usepackage{newfloat}
\usepackage{listings}
\usepackage{placeins}
\usepackage{caption}
\usepackage{subcaption}
\usepackage{adjustbox}

\usepackage{amsmath}
\usepackage{booktabs}

\usepackage{multirow}
\usepackage{newfloat}
\usepackage{listings}
\DeclareCaptionStyle{ruled}{labelfont=normalfont,labelsep=colon,strut=off} %
\floatstyle{ruled}
\newfloat{listing}{tb}{lst}{}
\floatname{listing}{Listing}
\title{Persuasion Strategies in Advertisements}
\author {
    Yaman Kumar\footnote{Contact Email: yamank@iiitd.ac.in, ykumar@adobe.com}\textsuperscript{\rm 1,2,3},
    Rajat Jha\textsuperscript{\rm 1},
    Arunim Gupta\textsuperscript{\rm 1},
    Milan Aggarwal\textsuperscript{\rm 2},\\
    Aditya Garg\textsuperscript{\rm 1},
    Tushar Malyan\textsuperscript{\rm 1}, Ayush Bhardwaj\textsuperscript{\rm 1},\\Rajiv Ratn Shah\textsuperscript{\rm 1}, Balaji Krishnamurthy\textsuperscript{\rm 2}, and Changyou Chen\textsuperscript{\rm 3}
}
\affiliations {
    \textsuperscript{\rm 1} IIIT-Delhi,\\ \textsuperscript{\rm 2} Adobe Media and Data Science Research (MDSR),\\ \textsuperscript{\rm 3} University at Buffalo\\
}

\usepackage{bibentry}

\begin{document}

\maketitle
\everypar{\looseness=-1}

\begin{abstract}
Modeling what makes an advertisement persuasive, \textit{i.e.}, eliciting the desired response from consumer, is critical to the study of propaganda, social psychology, and marketing. Despite its importance, computational modeling of persuasion in computer vision is still in its infancy, primarily due to the lack of benchmark datasets that can provide persuasion-strategy labels associated with ads. Motivated by persuasion literature in social psychology and marketing, we introduce an extensive vocabulary of persuasion strategies and build the first ad image corpus annotated with persuasion strategies. We then formulate the task of persuasion strategy prediction with multi-modal learning, where we design a multi-task attention fusion model that can leverage other ad-understanding tasks to predict persuasion strategies. %
The dataset also provides image segmentation masks, which labels persuasion strategies in the corresponding ad images on the test split. We publicly release our code and dataset at \url{https://midas-research.github.io/persuasion-advertisements/}.
\end{abstract}

\section{Introduction}
\begin{figure}
    \centering
    \includegraphics[scale=0.079]{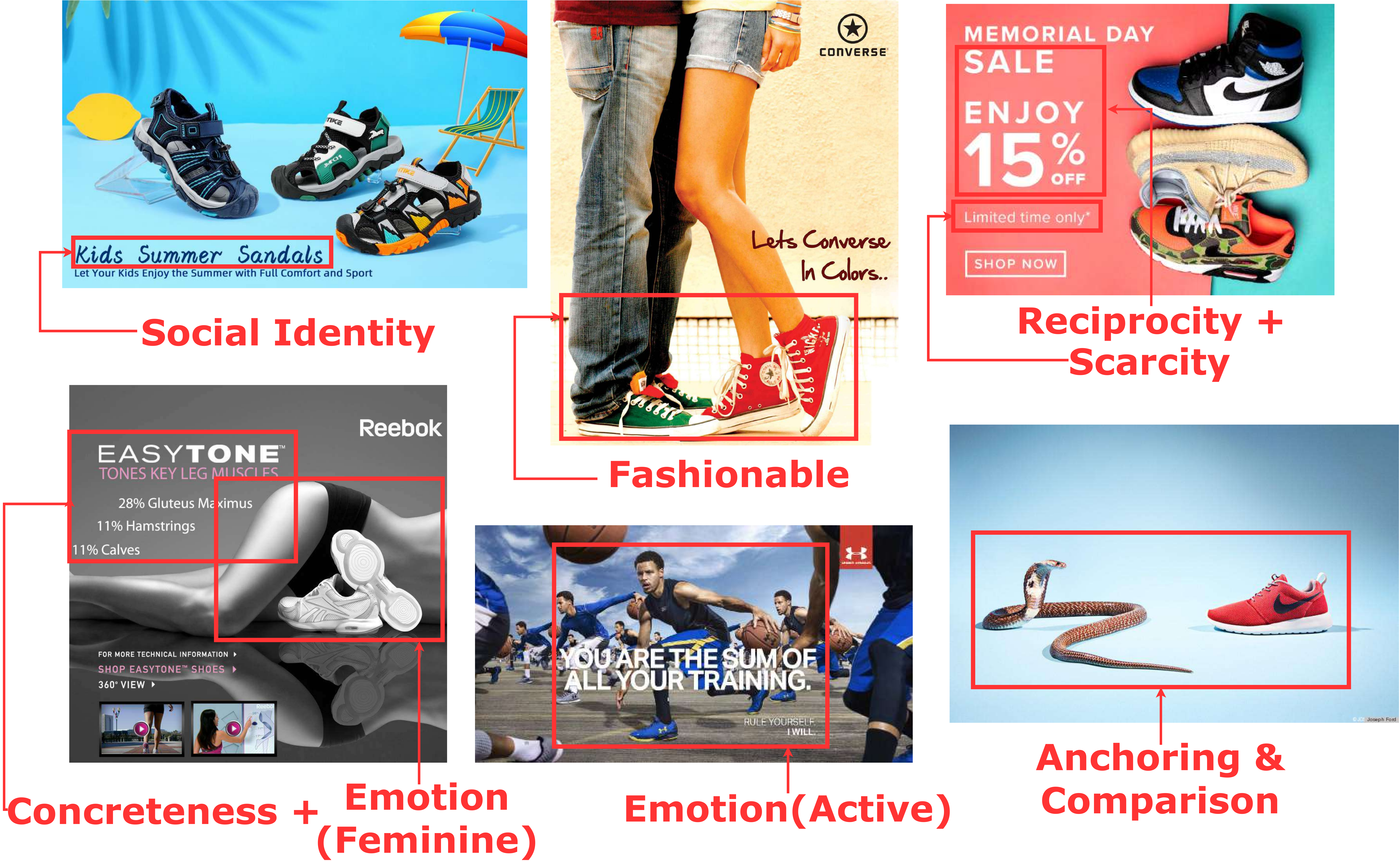}
    \caption{Different persuasion strategies are used for marketing the same product (footwear in this example). The strategies are in red words and to be defined by us in the paper.}
    \label{fig:footwear-strategies}
\end{figure}

Marketing communications is the mode by which companies and governments inform, remind, and persuade their consumers about the products they sell. They are the primary means of connecting brand with consumers through which the consumer can know what the product is about, what it stands for, who makes it, and can be motivated to try it out. %
To introduce meaning into their communication, marketers use various rhetorical devices in the form of persuasion strategies such as \textbf{emotions} ({\it e.g.}, Oreo's ``Celebrate the Kid Inside'', humor by showing Ronald McDonald sneaking into the competitor Burger King's store to buy a burger), \textbf{reasoning} ({\it e.g.}, ``One glass of Florida orange juice contains 75\% of your daily vitamin C needs''), \textbf{social identity} ({\it e.g.}, Old Spice's ``Smell like a Man''), and \textbf{impact} ({\it e.g.}, Airbnb showing a mother with her child with the headline ``My home is funding her future'')\footnote{Refer Appendix:Fig:1 
for seeing these ads.%
}. 
Similarly, even for marketing the same product, marketers use different persuasion strategies to target different demographies (see Fig.~\ref{fig:footwear-strategies}). %
Therefore, recognizing and understanding persuasion strategies in ad campaigns is vitally important to decipher viral marketing campaigns, propaganda, and enable ad-recommendation.

\begin{figure*}[h]
        \centering
        \includegraphics[scale=0.188]{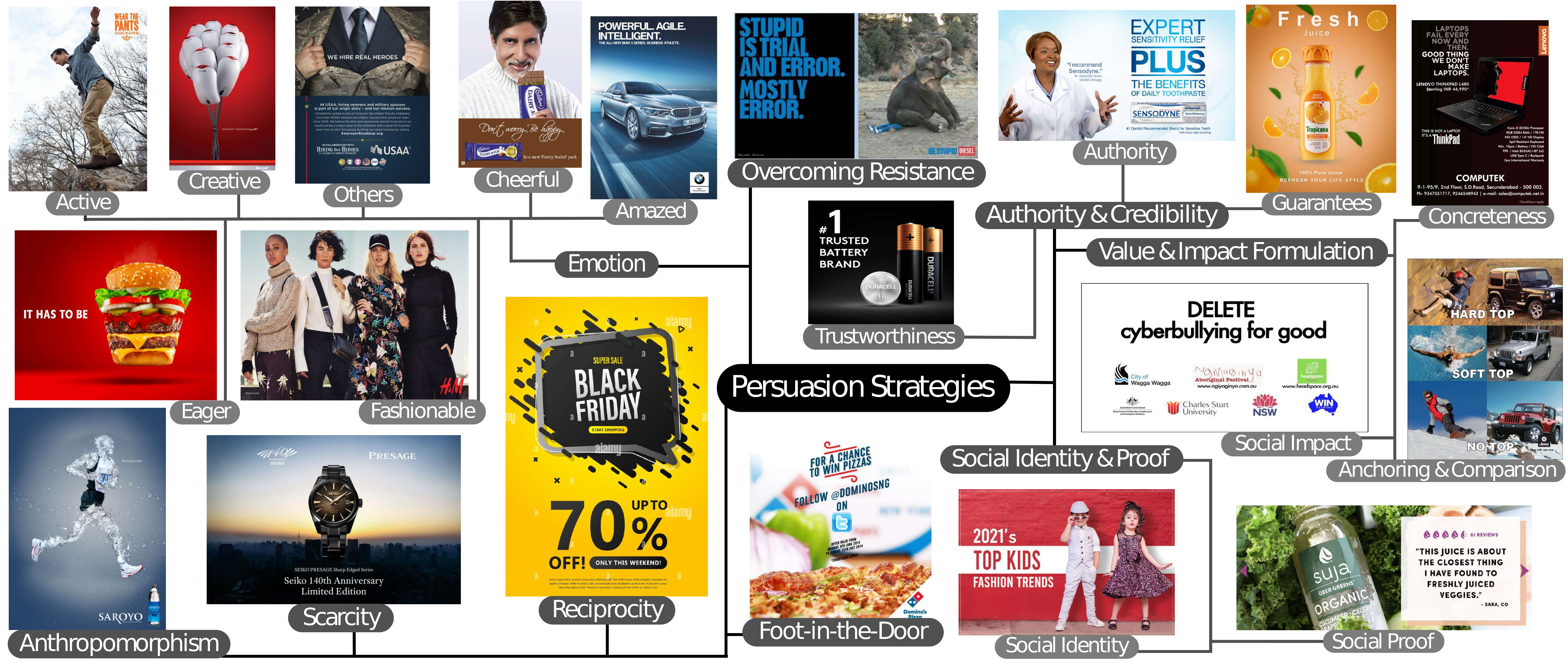}
        \caption{\looseness=-1 Persuasion strategies in advertisements. Marketers use both text and vision modalities to create ads containing different messaging strategies. Different persuasion strategies are constituted by using various rhetorical devices such as slogans, symbolism, colors, emotions, allusion.}
        \label{fig:persuasion-strategies-headline-image}
\end{figure*}

Studying rhetorics of this form of communication is an essential part of understanding visual communication in marketing. Aristotle, in his seminal work on rhetoric, underlining the importance of persuasion, equated studying rhetorics with the study of persuasion\footnote{``Rhetoric may be defined as the faculty of discovering in any particular case all of the available means of \textit{persuasion}'' \cite{rapp2002aristotle}} \cite{rapp2002aristotle}. While persuasion is studied extensively in social science fields, including marketing \cite{meyers1999consumers,keller2003affect} and psychology \cite{hovland1953communication,petty1986elaboration}, computational modeling of persuasion in computer vision is still in its infancy, primarily due to the lack of benchmark datasets that can provide representative corpus to facilitate this line of research. In the limited work that has happened on persuasion in computer vision, researchers have tried to address the question of which image is more persuasive \cite{bai2021m2p2} or extracted low-level features (such as emotion, gestures, and facial displays), which indirectly help in identifying persuasion strategies without explicitly extracting the strategies themselves \cite{joo2014visual}. On the other hand, decoding persuasion in textual content has been extensively studied in natural language processing from both extractive, and generative contexts \cite{habernal2016makes,ChenYang2021,luu2019measuring}. This forms the motivation of our work, where we aim to identify the persuasion strategies used in visual content such as advertisements.

The systematic study of persuasion began in the 1920s with the media-effects research by \citet{lasswell1971propaganda}, which was used as the basis for developing popular models of persuasion, like the Elaboration Likelihood Model (ELM) \cite{petty1986elaboration}, Heuristic Systematic Model (HSM) \cite{chaiken1980heuristic}, and Hovland's attitude change approach \cite{hovland1953communication}. %
These models of persuasion posit a dual process theory that explains attitude and behavior change (persuasion) in terms of the following major factors: stimuli (messages), personal motivation (the desire to process the message), capability of critical evaluation, and cognitive busyness. These factors could be divided into cognitive, behavioral, and affective processes of attitude change. 
In this work, we build on these psychological insights from persuasion models in sociology and marketing and study the message strategies that lead to persuasion. We codify, extend, and unify persuasion strategies studied in the psychology and marketing literature into a set of 20 strategies divided into 9 groups (see Fig.~\ref{fig:persuasion-strategies-headline-image}, Table~\ref{tab:persuasive-strategies-list}): \textit{Authority and Credibility}, \textit{Social Identity and Proof}, where cognitive indirection in the form of group decisioning and expert authority is used for decisions, \textit{Value and Impact Formulation} where logic is used to explain details and comparisons are made, \textit{Reciprocity}, \textit{Foot in the door}, \textit{Overcoming Resistance} where social and cognitive consistency norms are harnessed to aid decision-making, \textit{Scarcity}, \textit{Anthropomorphism} and \textit{Emotion} where information is evaluated from the lenses of feelings and emotions. In addition to introducing the most extensive vocabulary for persuasion strategies, we make a superset of persuasion strategies presented in the prior NLP works, which introduced text and domain-specific
persuasion tactics, thus making large-scale understanding of persuasion across multiple contexts comparable and replicable.

Constructing a large-scale dataset containing persuasion strategies labels is time-consuming and expensive. We leverage active learning to mitigate the cost of labeling fine-grained persuasion strategies in advertisements. We first introduce an attention-fusion model trained in a multi-task fashion over modalities such as text, image, and symbolism. We use the action-reason task from the Pitts Ads dataset \cite{hussain2017automatic} to train the model and then annotate the raw ad images from the same dataset for persuasion strategies based on an entropy based active learning technique.

To sum up, our contributions include:\\
1. We construct the largest set of generic persuasion strategies based on theoretical and empirical studies in marketing, social psychology, and machine learning literature. \\
2. We introduce the first dataset for studying persuasion strategies in advertisements. This enables initial progress on the challenging task of automatically understanding the messaging strategies conveyed through visual advertisements. We also construct a prototypical dataset containing image segmentation masks annotating persuasion strategies in different segments of an image.\\
3. We formulate the task of predicting persuasion strategies with a multi-task attention fusion model.\\
4. We conduct extensive experiments on the released corpus, showing the effect of different modalities on identifying persuasion strategies, correlation between strategies and topics and objects with different strategies. %

\begin{table*}[!h]
    \centering

    \FloatBarrier
{\fontsize{9.15}{10.15}\selectfont
\resizebox{1.0\textwidth}{!}{\begin{tabularx}{1.0\textwidth}[|l|]{>{\hsize=0.412\hsize\bfseries}X |>{\hsize=0.54\hsize\em} X | >{\hsize=1.56\hsize}X | >{\hsize=1.53\hsize}X}
    \hline \textbf{Group} & \textbf{Strategy} & \textbf{Definition} & \makecell{\textbf{Representative Prior Work}}\\\hline\hline
    
    \multirow{3}{*}[-2em]{\makecell{Authority\\and\\Credibility}} & Guarantees & Guarantees reduce risk and people try out such products more often. & \multirow{3}{=}{\looseness=-1\textbf{SPM}:\cite{aronson1963communicator,milgram1978obedience,cialdini2007influence,milgram1963behavioral,mcginnies1980better,giffin1967contribution,petty1986elaboration}\\\textbf{ML}:\cite{anand2011believe,iyer2019unsupervised,wachsmuth2017computational,ChenYang2021,durmus-cardie-2018-exploring}} \\
    & Authority & Authority indicated through expertise, source of power, third-party approval, credentials, and awards & \\
    & Trustworthiness & Trustworthiness indicated honesty and integrity of the source through tropes like years of experience, ``trusted brand'', numbers and statistics & \\\hline

    \multirow{2}{*}[-2em]{\makecell{Social\\Identity\\and\\Proof}} & Social Identity &
         \textit{Normative} influence, which involves conformity with the positive expectations of ``another'', who could be ``another person, a group, or one's self'' (includes self-persuasion, fleeting attraction, alter-casting, and exclusivity)
    & \multirow{2}{=}{\textbf{SPM}:\cite{deutsch1955study,petty1997attitudes,wood2000attitude,cialdini2004social,levesque2020human} \textbf{ML}: \cite{anand2011believe,iyer2019unsupervised,rosenthal2017detecting,yang2019let,zhang2016inferring}}\\
    
    & Social Proof & \textit{Informational influence} by accepting information obtained from others as evidence about reality, \textit{e.g.}, customer reviews and ratings &   \\\hline

         \multirow{1}{*}{Reciprocity} & Reciprocity & By \textit{obligating} the recipient of an act to repayment in the future, the rule for reciprocation begets a sense of future obligation, often unequal in nature & \looseness=-1\textbf{SPM}:\cite{regan1971effects,cialdini2007influence,clark1984record,clark1979interpersonal,clark1986keeping} \textbf{ML}:\cite{anand2011believe,iyer2019unsupervised,althoff2014ask,ChenYang2021,shaikh-etal-2020-examining} \\\hline

    \multirow{1}{*}{\makecell{Foot in\\the door}}
        & Foot in the door & Starting with small requests followed by larger requests to facilitate compliance while maintaining \textit{cognitive coherence}.& {\textbf{SPM}: \cite{freedman1966compliance,burger1999foot,cialdini2007influence} \textbf{ML}:\cite{chen2021weakly,wang-etal-2019-persuasion,vargheese2020exploring}}\\\hline

    \multirow{1}{*}{\makecell{Overcoming\\Resistance}} & \makecell{Overcoming\\Resistance} & Overcoming resistance (reactance) by postponing consequences to the future, by focusing resistance on realistic concerns, by forewarning that a message will be coming, by acknowledging resistance, by raising self-esteem and a sense of efficacy. & \textbf{SPM}:\cite{mcguire1961relative,knowles2004resistance,mcguire1964inducing}\newline\textbf{ML}:\{None\}\\\hline

     \multirow{3}{*}[-2em]{\makecell{Value and\\Impact\\Formulation}} & Concreteness & Using concrete facts, evidence, and statistics to appeal to the logic of consumers & \multirow{3}{=}{\textbf{SPM}:\cite{lee2010value,furnham2011literature,wegener2001implications,tversky1974judgment,strack1997explaining,bhattacharya2003consumer}%
    \hspace{1mm}\textbf{ML}:\cite{zhang2017characterizing,longpre2019persuasion}}\\
     
        & \makecell{Anchoring and\\Comparison} & A product's value is strongly influenced by what it is compared to. &\\
        
        & Social Impact & Emphasizes the importance or bigger (societal) impact of a product &\\\hline

    Scarcity & Scarcity & People assign more value to opportunities when they are less available. This happens due to psychological reactance of losing freedom of choice when things are less available or they use availability as a cognitive shortcut for gauging quality. & \textbf{SPM}:\cite{brehm1966theory,lynn1991scarcity,rothman1999systematic,tversky1985framing}\newline\textbf{ML}:\cite{yang2019let,ChenYang2021,shaikh-etal-2020-examining}\\\hline

    \multirow{1}{*}{\makecell{Anthropo-\\morphism}} & \makecell{Anthropo-\\morphism} & \makecell{When a brand or product is seen as human-\\like, people will like it more and feel closer\\to it.} & {\textbf{SPM}:\cite{fournier1998consumers,levesque2020human,epley2007seeing} \textbf{ML}:\{None\}} \\\hline

     \multirow{7}{*}{Emotion} & Amazed & \multirow{7}{=}{Aesthetics, feeling and other non-cognitively demanding features used for persuading consumers}& \multirow{7}{=}{\looseness=-1\textbf{SPM}:\cite{hibbert2007guilt,petty1986elaboration,petty1983central}\\\textbf{ML}:\cite{yang2019let,tan2016winning,hidey2017analyzing,he2018decoupling,durmus-cardie-2018-exploring,zhang2017characterizing,wachsmuth2017computational}}\\
        & Fashionable & & \\
        & Active,Eager & & \\
        & Feminine & & \\
        & Creative  & & \\
        & Cheerful  & & \\
        & Further Minor &  & \\\hline

    \makecell{Unclear} & \makecell{Unclear} & If the ad strategy is unclear or it is not in English & \\\hline

\end{tabularx}}}
\caption{The generic taxonomy of persuasive strategies, their definitions, examples, and connections with prior work. Representative literature from a)~\texttt{SPM}: Social Psychology and Marketing, b)~\texttt{ML}: Machine Learning
\label{tab:persuasive-strategies-list}
}
\end{table*}
\FloatBarrier

\section{Related Work}
\label{sec:related work}

\textit{How do messages change people's beliefs and actions?} The systematic study of persuasion has captured researchers’ interest since the advent of mass influence mechanisms such as radio, television, and advertising. Work in persuasion spans across multiple fields, including psychology, marketing, and machine learning.

\noindent \textbf{Persuasion in Marketing and Social Psychology:} Sociology and communication science have studied persuasion for centuries now, starting from the seminal work of Aristotle on rhetoric. Researchers have tried to construct and validate models of persuasion. Due to space constraints, while we cannot cover a complete list of literature, in Table~\ref{tab:persuasive-strategies-list}, we list the primary studies which originally identified the presence and effect of various persuasion tactics on persuadees. We build on almost a century of this research and crystallize them into the persuasion strategies we use for annotation and modeling. %
Any instance of (successful) persuasion is composed of two events: (a) an attempt by the persuader, which we term as the persuasion strategies, and (b) subsequent uptake and response by the persuadee \cite{anand2011believe,vakratsas1999advertising}. In this work, we study (a) only while leaving (b) for future work. Throughout the rest of the paper, when we say persuasion strategy, we mean the former without considering whether the persuasion was successful or not.

\noindent \textbf{Persuasion in Machine Learning:} Despite extensive work in social psychology and marketing on persuasion, most of the work is qualitative, where researchers have looked at a small set of messages with various persuasion strategies to determine their effect on participants. Computational modeling of persuasion is still largely lacking. In the limited work in computational modeling of persuasion, almost all of it is concentrated in the NLP literature, with only very few works in computer vision.
Research on persuasion in NLP under the umbrella of argumentation mining is broadly carried out from three perspectives: extracting persuasion tactics, studying the effect of constituent factors on persuasion, and measurement of persuasiveness nature of content. \citet{anand2011believe,stab2014annotating,tan2016winning,chen2021weakly} are some examples of research studies which annotate persuasive strategies in various forms of persuader-persuadee interactions like discussion forums, social media, blogs, academic essays, and debates. We use these and other studies listed in Table~\ref{tab:persuasive-strategies-list} to construct our vocabulary of persuasion strategies in advertisements. Other studies focus on factors such as argument ordering \cite{shaikh-etal-2020-examining,li-etal-2020-exploring-role}, target audience \cite{lukin-etal-2017-argument}, and prior beliefs \cite{el-baff-etal-2020-analyzing} for their effect in bringing about persuasion. Studies such as \citet{althoff2014ask,raifer2022designing,wei2016post} also try to measure persuasiveness and generate persuasive content.

As one of the first works in the limited work in the computer vision domain, \citet{joo2014visual} introduced syntactical and intent features such as facial displays, gestures, emotion, and personality, which result in persuasive images. Their analysis was done on human images, particularly politicians, during their campaigns. %
\citet{joo2014visual}'s work on political campaigners is more restrictive than general product and public-service advertisements. Moreover, they deal with low-level features such as gestures and personality traits depicted through the face, which are important for detecting persuasion strategies but are not persuasion strategies themselves. Recently, \citet{bai2021m2p2} studied persuasion in debate videos where they proposed two tasks: debate outcome prediction and intensity of persuasion prediction. Through these tasks, they predict the persuasiveness of a debate speech, which is orthogonal to the task of predicting the strategy used by the debater.%

\section{Persuasion Strategy Corpus Creation}
\label{sec:Persuasion Strategy Corpus Creation}

\begin{figure}[t]
        \centering
        \includegraphics[scale=0.24]{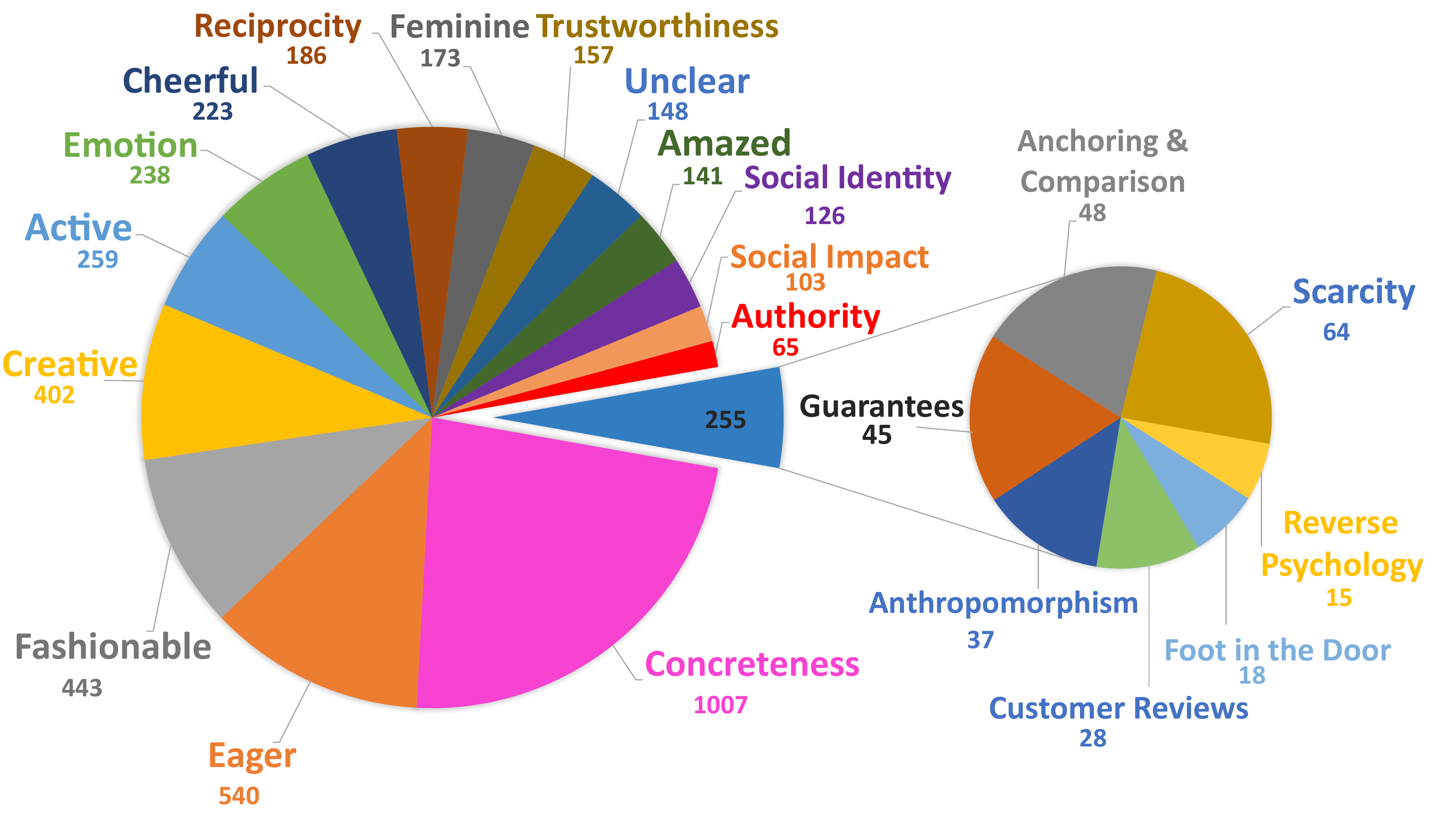}
        \caption{Distribution of Persuasion Strategies. The top-3 strategies are Concreteness, Eager, and Fashionable.}
        \label{fig:persuasion-strategies-distribution}
    \end{figure}

To annotate persuasion strategies on advertisements, we leverage raw images from the Pitts Ads dataset. It contains 64,832 image ads with labels of topics, sentiments, symbolic references (\textit{e.g.} dove symbolizing peace), and reasoning the ad provides to its viewers (see Appendix:Fig:2 
for a few examples). The dataset had ads spanning multiple industries, products, services, and also contained public service announcements. Through this, they presented an initial work for the task of understanding visual rhetoric in ads. Since the dataset already had a few types of labels associated with the ad images, we used active learning on a model trained in a multi-task learning fashion over the reasoning task introduced in their paper. We explain the model and then the annotation strategy followed in \S\ref{sec:Persuasion Strategy Prediction Model}.

To commence training, we initially annotated a batch of $250$ ad images with persuasion strategies defined in Table~\ref{tab:persuasive-strategies-list}. We recruited four research assistants to label persuasion strategies for each advertisement. Definitions and examples of different persuasion strategies were provided, together with a training session where we asked annotators to annotate a number of example images and walked them through any disagreed annotations. To assess the reliability of the annotated labels, we then asked them to annotate the same 500 images and computed Cohen's Kappa statistic to measure inter-rater reliability. We obtained an average score of 0.55. The theoretical maximum of Kappa given the unequal distribution is 0.76. In such cases, \citet{cohen1960coefficient} suggested that one should divide kappa by its maximum value $\mathbf{k}/\mathbf{k_{max}}$, which comes out to be 0.72. This is substantial agreement. Further, to maintain the labeling consistency, each image was double annotated, with all discrepancies resolved by an intervention of the third annotator using a majority voting.

The assistants were asked to label each image with no more than 3 strategies. If an image had more than 3 strategies, they were asked to list the top-3 strategies according to the area covered by the pixels depicting that strategy. In total, we label 3000 ad-images with their persuasion strategies; and the numbers of samples in train, val and test split are 2500, 250 and 250 resp.\footnote{Appendix:Table:1 
shows the detailed distribution of number of strategies in ads} Fig.~\ref{fig:persuasion-strategies-distribution} presents the distribution of persuasion strategies in the dataset. It is observed that concreteness is the most used strategy in the dataset, followed by eager and fashionable. The average number of strategies in an ad is 1.49, and the standard deviation is 0.592. We find that scarcity (92.2\%), guarantees (91.1\%), reciprocity (84.4\%), social identity (83.3\%), and cheerful (83\%), %
are the top 5 strategies, which occur in groups of 2 or 3. We observe that the co-occurrence of these strategies is due to the fact that many of them cover only a single modality (\textit{i.e.}, text or visual), leaving the other modality free for a different strategy. For example, concreteness is often indicated by illustrating points in text, while the visual modality is free for depicting, say, emotion. See Fig:3 %
in the Appendix for an example, where the image depicting concreteness also has the social impact strategy in it. Similarly, feminine emotion is also depicted in Fig.~\ref{fig:footwear-strategies}, along with concreteness.

\begin{figure}[t]
    \centering
    \begin{subfigure}[b]{0.45\columnwidth}
        \includegraphics[scale=0.22]{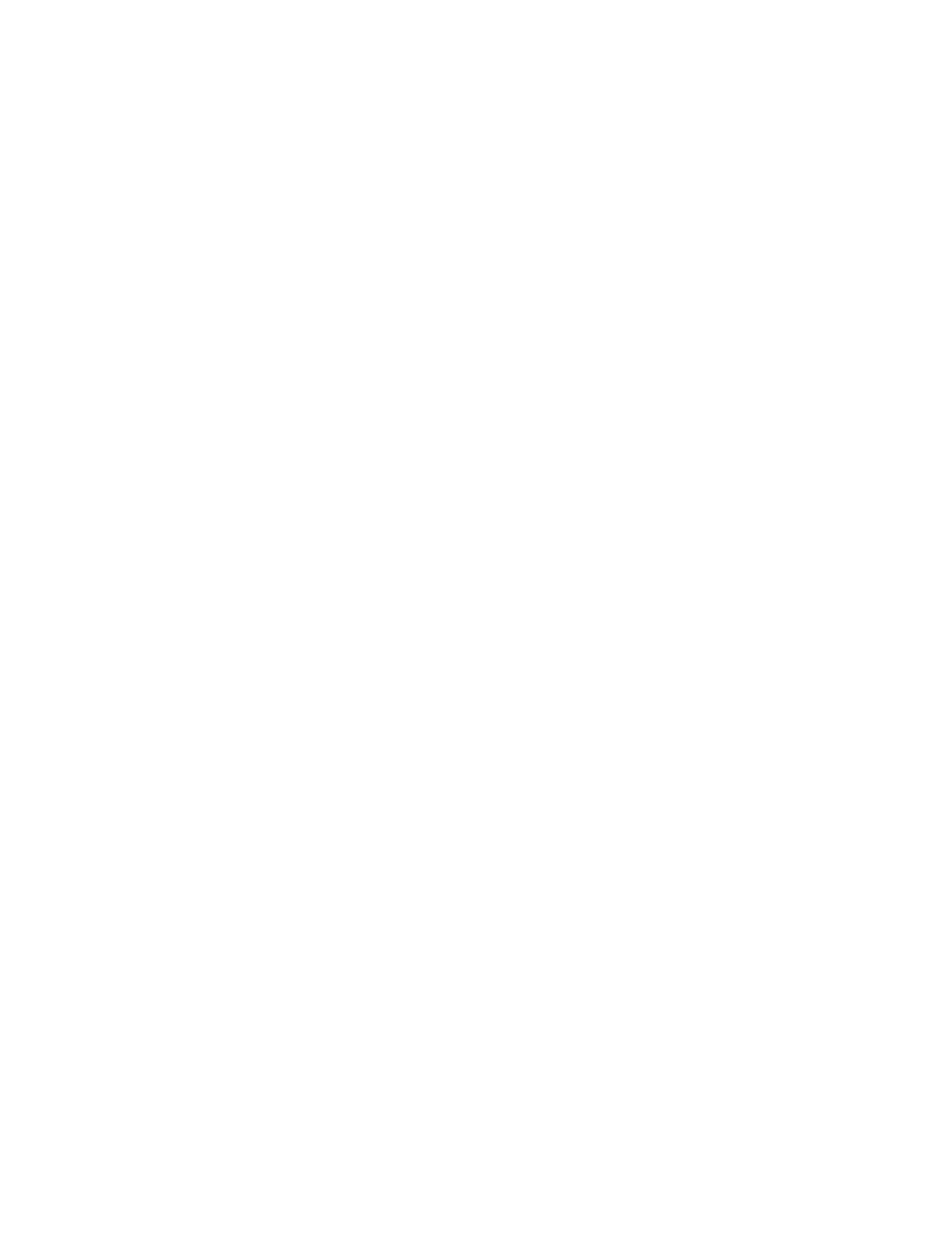}
    \end{subfigure}
    \begin{subfigure}[b]{0.45\columnwidth}
        \includegraphics[scale=0.22]{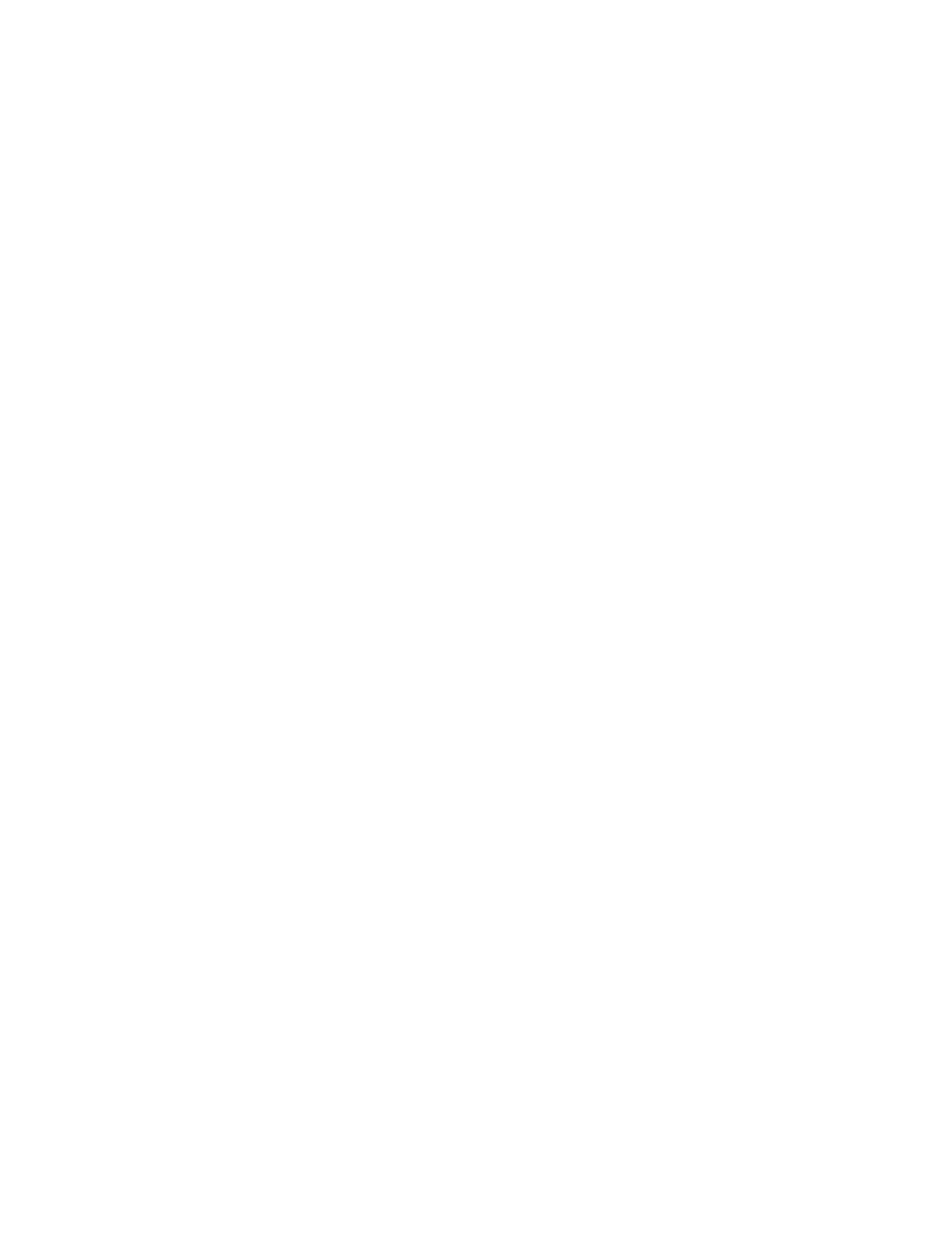}
    \end{subfigure}
    \caption{\label{fig:segmented image} Image with a segmentation mask depicting the strategies \textit{Emotion:Cheerful}, \textit{Emotion:Eager} and \textit{Trustworthiness}. Appendix:Fig:4 %
    contains more such examples.}
\end{figure}

Next, we calculate the Dice correlation coefficient statistics for pairs of co-occurring persuasion strategies. The top-5 pairs are eager-concreteness (0.27), scarcity-reciprocity (0.25), eager-cheerful (0.19), amazed-concreteness (0.17), and eager-reciprocity (0.17). We find that these correlation values are not particularly high since marketers seldom use \textit{common pairings} of messaging strategies to market their products. The visual part mostly shows eager strategy in ads; therefore, we find that the text modality becomes free to show other strategies. That is why primarily text-based concreteness, cheerfulness, and reciprocity strategies are present in the text modality with the visual-based eager strategy. On the other hand, primarily vision-based amazement, eagerness, and scarcity (short-text) strategies co-occur with text-based reciprocity and concreteness (E.g., see Fig.~\ref{fig:footwear-strategies}).

Next, we calculate the correlation between image topics and objects present with persuasion strategies. We see that the feminine and fashionable emotion strategies are most often associated with beauty products and cosmetics (corr=0.426, 0.289). This is understandable since most beauty products are aimed at women. We see that the fast-food and restaurant industries often use eagerness as their messaging strategy (corr = 0.588,0.347). We find that the presence of humans in ads is correlated with the concreteness strategy\footnote{see Appendix:Fig:3  %
for a few examples} (corr=0.383). On the other hand, vehicle ads use emotion:amazed and concreteness (corr=0.521,0.241)\footnote{See Appendix:Fig:5 
for detailed correlations.}. 
Similar to a low correlation in co-occurring strategies, we find that product segments and their strategies are not highly correlated. This is because marketers use different strategies to market their products even within a product segment. Fig.~\ref{fig:footwear-strategies} shows an example in which the footwear industry (which is a subsegment of the apparel industry) uses different strategies to market its products. Further, for a batch of $250$ images, we also label segmented image regions corresponding to the strategies present in the image. These segment masks were also double-annotated. Fig.~\ref{fig:segmented image} presents an example of masks depicting parts of the image masked with different persuasion strategies in a drink advertisement.

\section{Modeling: Persuasion Strategy Prediction}
\label{sec:Persuasion Strategy Prediction Model}

\begin{figure*}[h]
        \centering
        \includegraphics[width=1.0\textwidth]{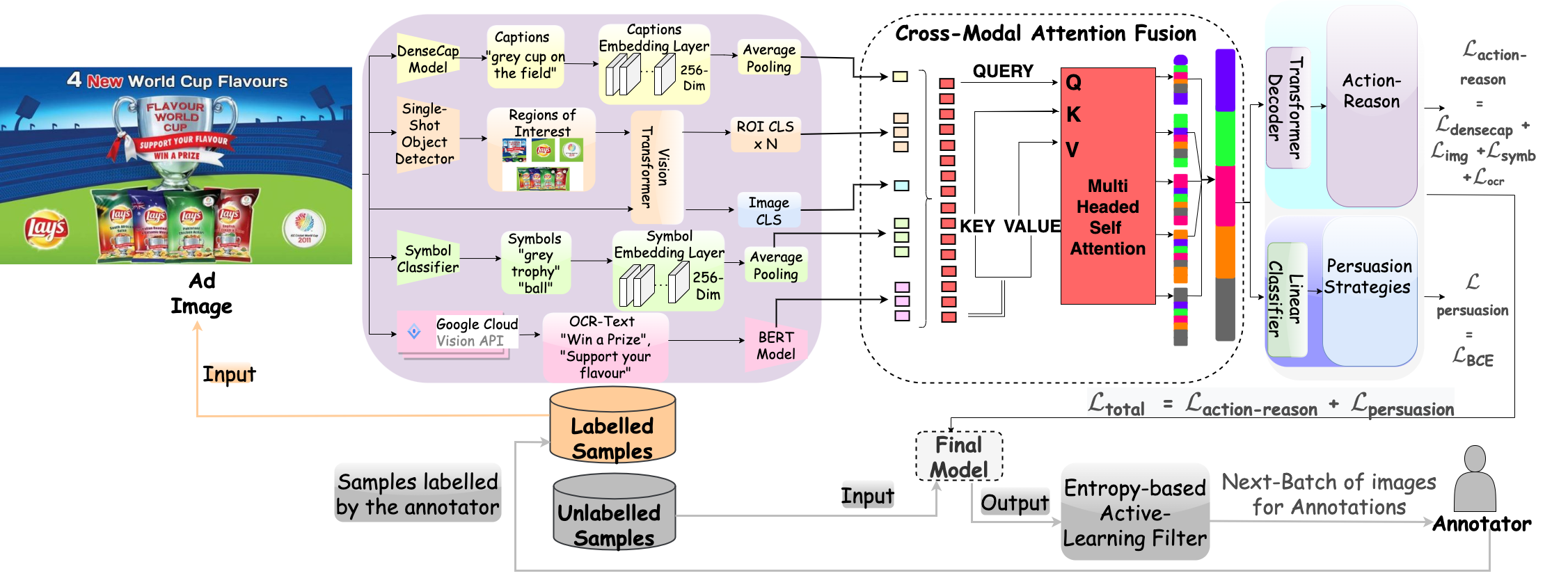}
        \caption{Architecture of the Persuasion Strategy Prediction model. To capture the different rhetoric devices, we extract features for the image, text, and symbolism modalities and then apply cross-modal attention fusion to leverage the interdependence of the different devices. Further, the model trains over two tasks: persuasion strategies and the reasoning task of action-reason prediction.}
\label{fig:arch_diag}
\end{figure*}

The proposed Ads dataset $\mathcal{D}$ annotated with the persuasion strategies comprises of samples where each sample advertisement $a_i$ is annotated with a set of annotation strategies $S_i$ such that $1\le|{S_i}|\le3$. The unique set of the proposed persuasion strategies $\mathcal{P}$ is defined in Table~\ref{tab:persuasive-strategies-list}. Given $a_i$, the task of the modeling is to predict the persuasion strategies present in the input ad. As we observe from Fig.~\ref{fig:persuasion-strategies-headline-image}, advertisements use various rhetoric devices to form their messaging strategy. The strategies thus are in the form of multi-modalities, including images, text and symbolism. To jointly model the modalities, we design an attention fusion multi-modal framework, which fuses multimodal features extracted from the ad, {\it e.g.}, the ad image, text present in the ad extracted through the OCR (Optical Character Recognition), regions of interest (ROIs) extracted using an object detector, and embeddings of captions obtained through an image captioning model (see Fig.~\ref{fig:arch_diag}). The information obtained through these modalities are firstly embedded independently through their modality specific encoders followed by a transformer-based cross-attention module to fuse the extracted features from different modalities. The fused embeddings from the attention module are then used as input for a classifier that predicts a probability score for each strategy $p \in \mathcal{P}$. The overall architecture of the proposed model is illustrated in Fig.\ref{fig:arch_diag}. In the following, we describe each step in the prediction pipeline in detail.

\subsection{Feature Extractors} 

In order to capture different rhetoric devices, we extract features from the image, text, and symbolism modalities.

\textbf{Image Feature:} We use the Vision Transformer \cite{dosovitskiy2020image} (ViT) model for extracting image features from the entire input image. The model resizes the input image to size $224 \times 224$ and divides it into patches of size $16 \times 16$. The model used has been pre-trained on the ImageNet 21k dataset. We only use the first output embedding, which is the CLS token embedding, a $768$ dimension tensor, as we only need a representation of the entire image. Then, a fully connected layer is used to reduce the size of the embedding, resulting in a tensor of dimension $256$.

\textbf{Regions of Interest (RoIs) from Detected Objects and Captions:} %
Ad images contain elements that the creator deliberately chooses to create \textit{intentional impact} and deliver some \textit{message} in addition to the ones that occur \textit{naturally} in the environment. Therefore, it is important to identify the composing elements of an advertisement to understand the creator's intention and the ad's message to the viewer. We detect and extract objects as regions of interest (RoIs) from the advertisement images. We get the RoIs by training the single-shot object detector model by \citet{liu2016ssd} on the COCO dataset \cite{lin2014microsoft}. We compare it with the recent YOLOv5 model \cite{redmon2016you}. We also extract caption embeddings to detect the most important activity from the image using a caption generation mode. We compare DenseCap \cite{yang2017dense} and the more recent BLIP \cite{li2022blip} for caption generation.

\textbf{OCR Text:} The text present in an ad presents valuable information about the brand, such as product details, statistics, reasons to buy the product, and creative information in the form of slogans and jingles that the company wants its customers to remember and thus making it helpful in decoding various persuasion strategies. Therefore, we extract the text from the ads and use it as a feature in our model. We use the Google Cloud Vision API for this purpose. All the extracted text is concatenated, and the size is restricted to $100$ words. We pass the text through a BERT model \cite{devlin2018bert} and use the final CLS embedding as our OCR features. Similar to image embeddings, an FC layer is used to convert embeddings to $256$ dimensional vectors. The final embedding of the OCR is a tensor of dimension $100 \times 256$.

\textbf{Symbolism:} While the names of the detected objects convey the names or literal meaning of the objects, creative images often also use objects for their symbolic and figurative meanings. For example, an upward-going arrow represents growth or the north direction or movement towards the upward direction depending on the context; similarly, a person with both hands pointing upward could mean danger (\textit{e.g.}, when a gun is pointed) or joy (\textit{e.g.}, during dancing). In Fig.~\ref{fig:persuasion-strategies-headline-image}, in the creative Microsoft ad, a symbol of a balloon is created by grouping multiple mice together. Therefore, we generate symbol embeddings to capture the symbolism behind the most prominent visual objects present in an ad. We use the symbol classifier by \citet{hussain2017automatic} on ad images to find the distribution of the symbolic elements present and then convert this to a $256$ dimension tensor. %

\subsection{Cross-Modal Attention} To capture the inter-dependency of multiple modalities for richer embeddings, we apply a cross-modal attention (CMA) layer~\cite{frank2021vision} to the features extracted in the previous steps. Cross-modal attention is a fusion mechanism where the attention masks from one modality (\textit{e.g.} text) are used to highlight the extracted features in another modality (\textit{e.g.} symbolism). It helps to link and extract common features in two or more modalities since common elements exist across multiple modalities, which complete and reinforce the message conveyed in the ad. For example, the pictures of the silver cup, stadium, and ball, words like ``Australian'', ``Pakistani'', and ``World Cup'' present in the chips ad shown in Fig.~\ref{fig:arch_diag} link the idea of buying \textit{Lays} with supporting one's country's team in the World Cup. Cross attention can also generate effective representations in the case of missing or noisy data or annotations in one or more modalities \cite{frank2021vision}. This is helpful in our case since marketing data often uses implicit associations and relations to convey meaning.

\begin{table}[t]
        \centering
        \resizebox{0.85\columnwidth}{!}{%
       \begin{tabular}{lll}%
             \textbf{\thead{Models}} & \textbf{\thead{Top-1  Acc.}} & \textbf{\thead{Top-3 Acc.}}\\ \midrule
             Our Model & \textbf{59.2} & \textbf{84.8} \\ \midrule
             w/o DenseCap & 55.6 & 80.8 \\ \midrule
             w/o Symbol   & 58.8 & 81.6 \\ \midrule
           w/o DenseCap \& Symbol & 55.2 & 80.8 \\ \midrule
            w/o OCR & 54.8 & 82 \\ \midrule
            w/o Symbol,\\OCR \& DenseCap & 58 & 78.8 \\ \midrule
            w/o Action-Reason Task & 56.4 & 80.4\\ \midrule
            Random Guess & 6.25 &  18.75\\\bottomrule
        \end{tabular}}
        \caption{Effect of different Modalities and Tasks on the accuracy and performance of the strategy prediction task.}
        \label{tab:modalities}
    \end{table}

The input to the cross-modal attention layer is constructed by concatenating the image, RoI, OCR, caption, and symbol embeddings. This results in a $114 \times 256$ dimension input to our attention layer. The cross-modal attention consists of two layers of transformer encoders with a hidden dimension size of $256$. %
The output of the attention layer gives us the final combined embedding of our input ad. 
Given image embeddings $E_i$, RoI embeddings $E_r$, OCR embeddings $E_o$, caption embeddings $E_c$ and symbol embeddings $E_s$, the output of the cross-attention layer $E_{att}$ is formulated as:
\begin{align}
        \text{Enc}(X) &= \text{CMA}(\left[E_i(X),E_r(X), E_o(X),E_c(X),E_s(X)\right])\, \nonumber
\end{align}
, where $[\ldots,\ldots]$ is the concatenation operation.

\subsection{Persuasion Strategy Predictor} This module is a persuasion strategy predictor, which processes the set of feature embedding \text{Enc}(X) obtained through cross-modality fusion. Specifically, \text{Enc}(X) is passed through a self-attention layer as:
\begin{align}
o_{1} = softmax(\text{Enc}(X) \otimes W_{self-attn})^\intercal \otimes \text{Enc}(X)
\end{align}
where \text{Enc}(X) is of the dimension $114\times256$,  $W_{self-attn} \in \mathcal{R}^{256 \times 1}$, $\otimes$ denote tensor multiplication and $o_1$ denotes the output of self attention layer, which is further processed through a linear layer to obtain $o_{|\mathcal{P}|}$ to represent the logits for each persuasion strategy. We apply sigmoid over each output logit such that the $i^{th}$ index of the vector after applying sigmoid denotes $p_i$ - the probability with which $i^{th}$ persuasion strategy is present in the ad image. Our choice of using sigmoid over softmax is motivated by the fact that multiple persuasion strategies can be present simultaneously in an ad image. Consequently, the entire model is trained in an end-to-end manner using binary cross-entropy loss $\mathcal{L}_{s}$ over logit for each strategy:
\begin{align}
        \mathcal{L}_{s} = \left[-y_i \log(p_i) - (1-y_i) \log(1 - p_i) \right]
\end{align}
where, $y_i$ is 1 if $i^{th}$ persuasion strategy is present in the ad and 0 otherwise. It can be observed in Table~\ref{tab:modalities} that our model achieves an accuracy of $59.2\%$, where a correct match is considered if the strategy predicted by the model is present in the set of annotated strategies for a given ad. Further, we perform several ablations where we exclude each modality while retaining all the other modalities. We note that for each modality, excluding the modality results in a noticeable decrease in accuracy, with significant decreases observed when excluding DenseCap ($\sim3.6\%$) and OCR ($\sim4.4\%$). Further, we observe that using DenseCap for obtaining caption embeddings, and SSD for object detection works better than BLIP and YOLOv5, respectively (see Table~\ref{tab:varying_cap_obj}). We also explore using focal loss~\cite{lin2017focal} in place of cross-entropy loss to handle class imbalance but observed that it led to degradation instead of improvements (top-1 acc. of $56.4\%$ \textit{vs} $59.2\%$ using cross-entropy).

\subsection{Multi Task Learning}
One of the key opportunities for our persuasion strategies data labeling and modeling task was the presence of additional labels already given in the base Pitts Ads dataset. In that, \citet{hussain2017automatic} had given labels about the reasoning task. For the reasoning task, the annotators were asked to provide answers in
the form ``I should [Action] because [Reason]." for each ad. In other words, they asked the annotators to describe \textit{what the viewer should do and why}, according to
the ad. Similar to the reasoning task, persuasion strategies provide various cognitive, behavioral, and affective reasons to try to elicit the motivation of the ad viewers towards their products or services. Therefore, we hypothesize that these natural language descriptions of \textit{why the viewers should follow} the ad will be informative in inferring the ad's persuasion strategy.

We formulate obtaining action-reason statement as a sequence generation task where the model learns to generate a sentence $Y^g=(y^g_1, \ldots, y^g_T)$ of length $T$ conditioned on advertisement $X$ by generating the sequence of tokens present in the action-reason statement. To achieve this, we use a transformer decoder module that attends on the features $\text{Enc}(X)$ as shown in Fig.~\ref{fig:arch_diag}. The annotated action-reason statement is used to train the transformer decoder as an auxiliary task to strategy prediction through the standard teacher forcing technique used in Seq2Seq framework. Please refer to Appendix for more architectural details about the action-reason generation branch. As shown in Table~\ref{tab:modalities}, generating action-reason as an auxiliary task improves the strategy prediction accuracy by $2.8\%$.

\begin{table}[!t]
        \centering

        \resizebox{0.85\columnwidth}{!}{%
       \begin{tabular}{llllll}%
                \textbf{\thead{Model Used}} & \textbf{\thead{Top-1 \\Accuracy}} & \textbf{\thead{Top-3 \\Accuracy}} &  \textbf{Recall} \\ \midrule
                 Model with \\DenseCap \& SSD & 59.2 & 84.8 & 74.59\\ \midrule
                 Model with \\BLIP \& YOLOv5 & 58.4 &  83.8 & 71.58 \\ \midrule
        \end{tabular}}
        \caption{Comparison of caption and object detection models. We noticed that BLIP while being more recent and trained on a larger dataset, generates more informatory captions for background objects which DenseCap successfully ignores.}
        \label{tab:varying_cap_obj}
    \end{table}

\subsection{Active Learning}
We use an active learning method to ease the large-scale label dependence when constructing the dataset. As in every active learning setting, our goal is to develop a learner that selects samples from unlabeled sets to be annotated by an oracle. Similar to traditional active learners \cite{lewis1994heterogeneous}, we use uncertainty sampling to perform the sample selection. In doing so, such function learns to score the unlabeled samples based on the expected performance gain they are likely to produce if annotated and used to update the current version of the localization model being trained. To evaluate each learner, we measure the performance improvements, assessed on a labeled test set at different training dataset sizes.

At every learning step $t$, a set of labeled samples $L_t$ is first used to train a model
$f_t$. Then, from an unlabeled pool $U_t=D-L_t$, an image instance $a$ is chosen by a selection function $g$. Afterwards, an oracle provides temporal ground-truth for the selected instance, and the labeled set $L_t$ is augmented with this new annotation. This process repeats until the desired performance is reached or the set $U_t$ is empty.

        \begin{figure}[t]
        \centering
        \includegraphics[scale=0.105]{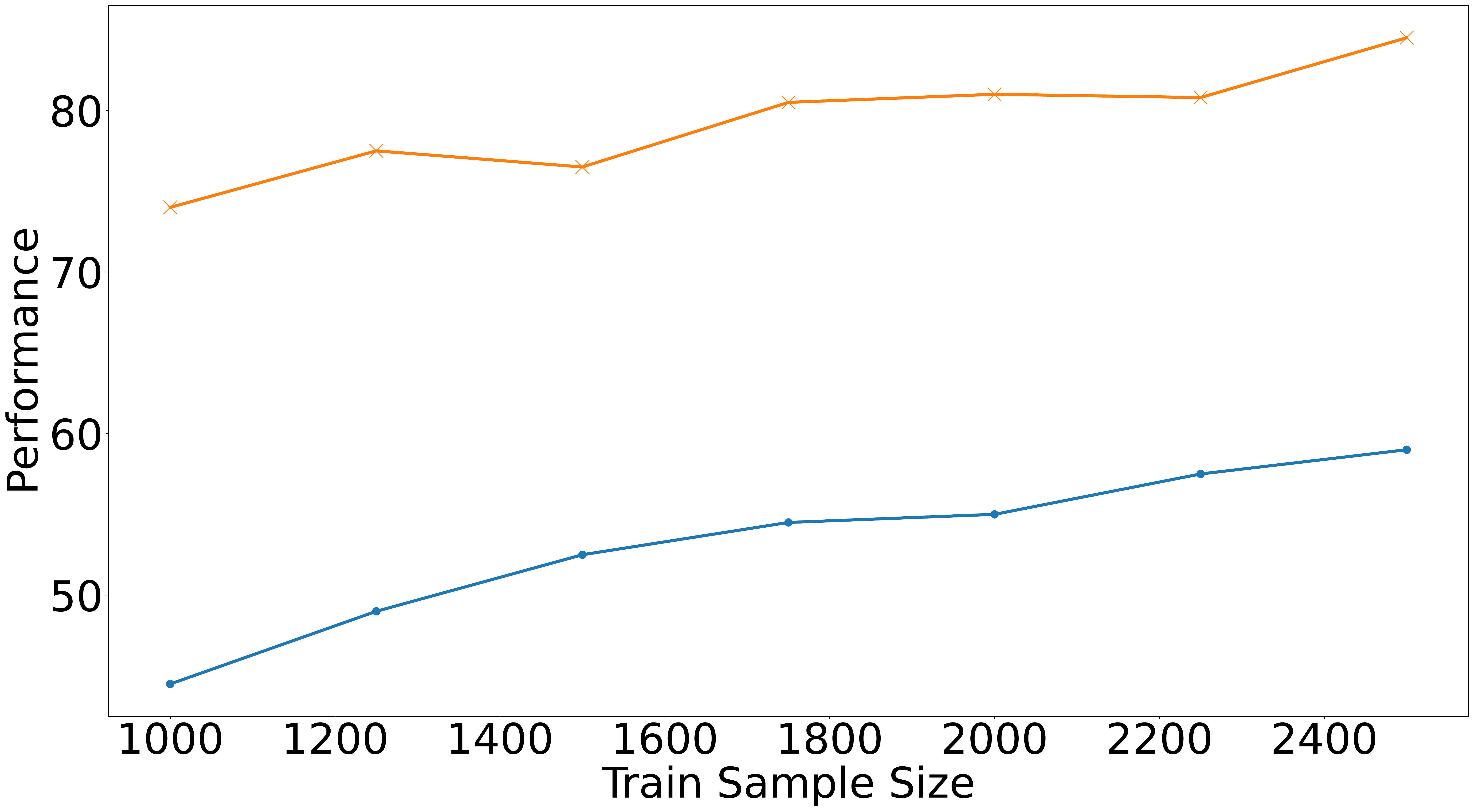}
        \caption{Incremental effect of introducing new data through active learning; Results for prediction of persuasion strategies on the test set}
        \label{fig:AL_batch_test}
    \end{figure}
In our implementation, we instantiate the active learning selection function as the entropy of the probability distribution predicted by the model over the set of persuasion strategies for a given ad image instance $a$. Formally, $g = -\sum_{i=1}^{|\mathcal{P}|}p^{n}_{i}*\log(p^{n}_i)$, where $p^{n}_i$ denotes the normalized probability with which $i^{th}$ persuasion strategy is present in $a$ as per the model prediction. The normalized probability $p^{n}_{i}$ is estimated as $p_i/\sum_{j=1}^{|\mathcal{P}|}p_j$. Intuitively, ad samples with high entropy selection values indicate that the model trained on limited data has a higher degree of confusion while predicting the persuasion strategy since it is not decisively confident about predicting few strategies. Hence, we rank the unlabeled ad images in the decreasing order of difficulty according to the corresponding values of the entropy selection function and select the top-k ads in the subsequent batch for annotation followed by training. We set k to 250 and analyze the effect of incrementally introducing new samples selected through active learning (Fig.~\ref{fig:AL_batch_test}). It can be seen that both top-1 and top-3 accuracy increases with the addition of new training data. We stop at the point when 2500 training samples are used since the model performs reasonably well with a top-1 and top-3 strategy prediction accuracy of 59.2\% and 84.8\% (see Fig.~\ref{fig:AL_batch_test}).

\section{Conclusion and Future Work}
\textit{What does an advertisement say which makes people change their beliefs and actions?} With limited works, the computational study of rhetoric of this all-pervasive form of marketing communication is still in its infancy. In this paper, based on the well-developed social psychology and marketing literature, we develop and release the largest vocabulary of persuasion strategies and labeled dataset. We develop a multi-task attention-fusion model for predicting strategies%
Future work can investigate how audience factors affect persuasion strategies across sectors. We would like to investigate what role do strategies play in viral marketing and how to generate advertisements given brands and strategies.
\section*{Acknowledgments}
We would like to thank Professor Diyi Yang, Georgia Institute of Technology for helping us with formulating and verifying the vocabulary of persuasion strategies and annotation guidelines.
{\small
\bibliography{arxiv-bib}
}

\clearpage
\setcounter{figure}{0}
\setcounter{table}{0}
\setcounter{section}{0}

\section*{Appendix}

\section{Definitions}
\textit{Dice Coefficient} : \; \(2*|X \cap Y|\: / \:(|X|+|Y|)\), \\
where X and Y are two sets;
a set with vertical bars on either side refers to the cardinality of the set, i.e. the number of elements in that set; and 
\(\cap\)  refers to the intersection of two sets.\\

\textit{Top-1 Accuracy} : It is defined as the fraction of images, where the highest predicted strategy is present in the ground-truth strategies.\\

\textit{Top-3 Accuracy} : It is defined as the fraction of images, where any of the top-3 highest predicted strategies is present in the ground-truth strategies.\\

\begin{table}[!h]
    \small
        \centering
        \resizebox{\columnwidth}{!}{%
       \begin{tabular}{llllll}%
       \toprule
             & \textbf{\thead{\#ads with \\1 strategy}} & \textbf{\thead{\#ads with \\2 strategies}} & \textbf{\thead{\#ads with \\3 strategies}} & \textbf{\thead{Avg.\\strategies}} & \textbf{\thead{Std.\\Dev.}}\\ \midrule
             Train-Set & 1440 & 905 & 155 & 1.486 & 0.612 \\\midrule
             Val-Set & 132 & 98 & 20 & 1.552  & 0.639 \\\midrule
             Test-Set & 147 & 93 & 10 & 1.452 &0.574\\\midrule\midrule
             Total & 1719 & 1096 & 185 & 1.49 & 0.592 \\\bottomrule
        \end{tabular}}
        \caption{\small Distribution of test, train, validation, and the total dataset}
        \label{table:strategy-train-test-val-stats}
    \end{table}

\begin{figure*}[!hb]
    \centering
    \begin{subfigure}[b]{0.22\textwidth}
         \centering
         \includegraphics[width=\textwidth,scale=0.6]{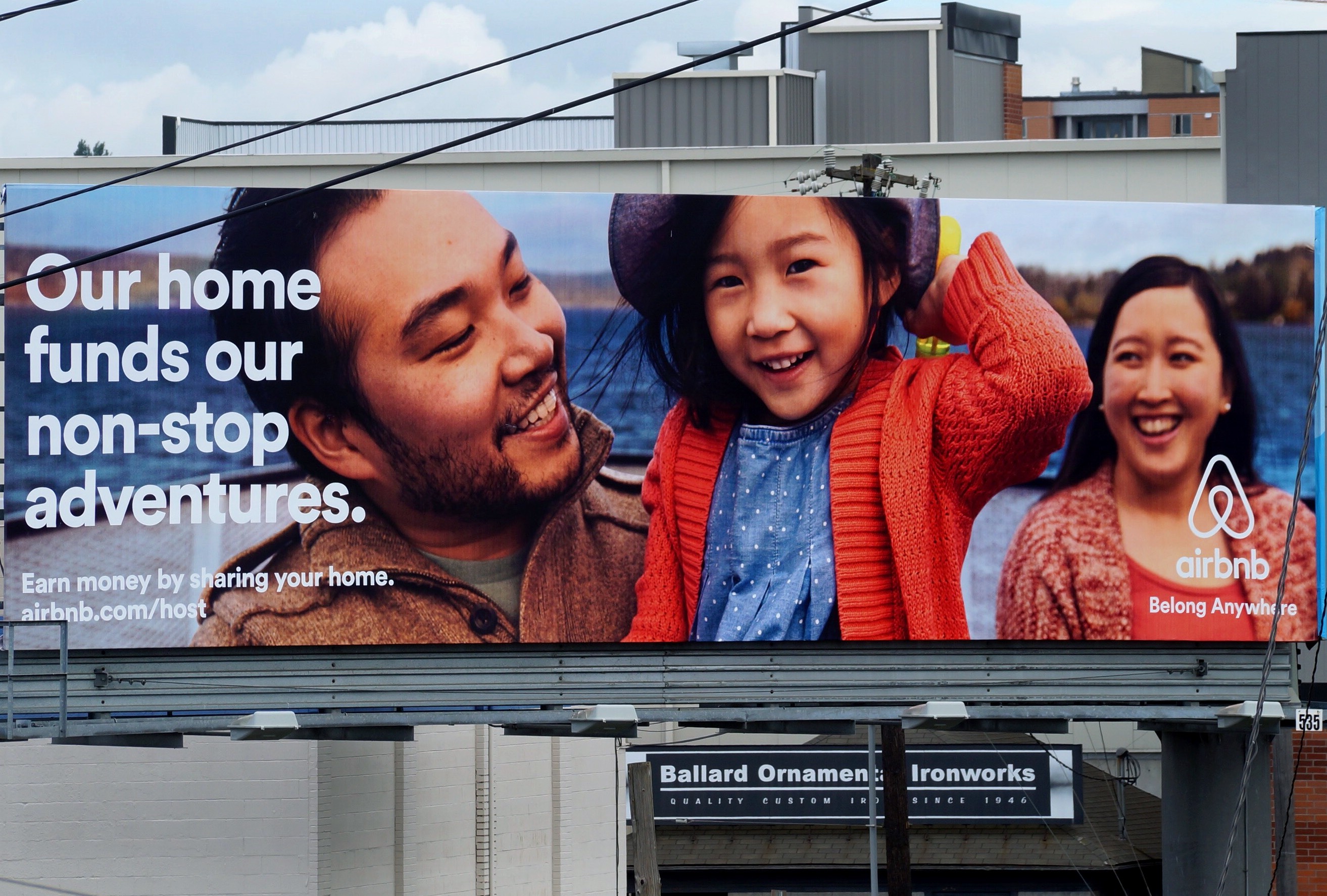}
         \caption{}
         \label{fig:Airbib}
     \end{subfigure}
     \begin{subfigure}[b]{0.2\textwidth}
         \centering
         \includegraphics[width=\textwidth,scale=0.5]{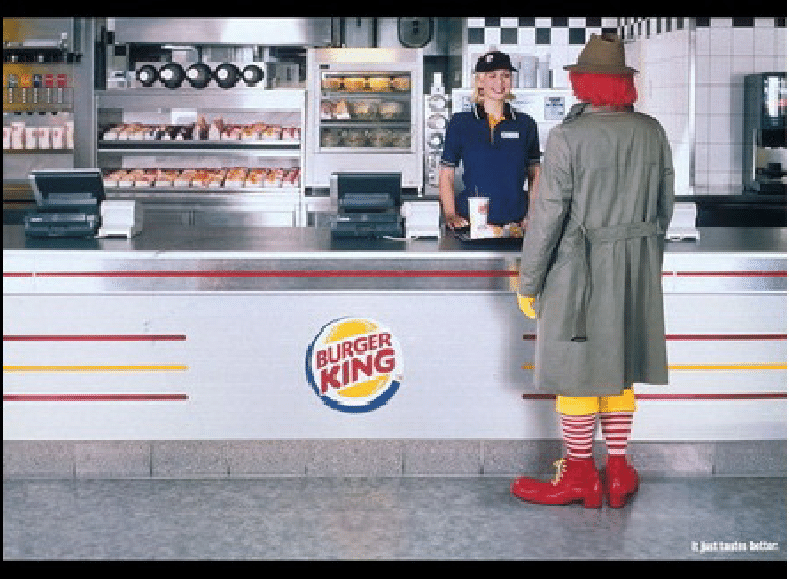}
         \caption{}
         \label{fig:mcdonald}
     \end{subfigure}
     \begin{subfigure}[b]{0.27\textwidth}
         \centering
         \includegraphics[width=\textwidth,scale=1.0]{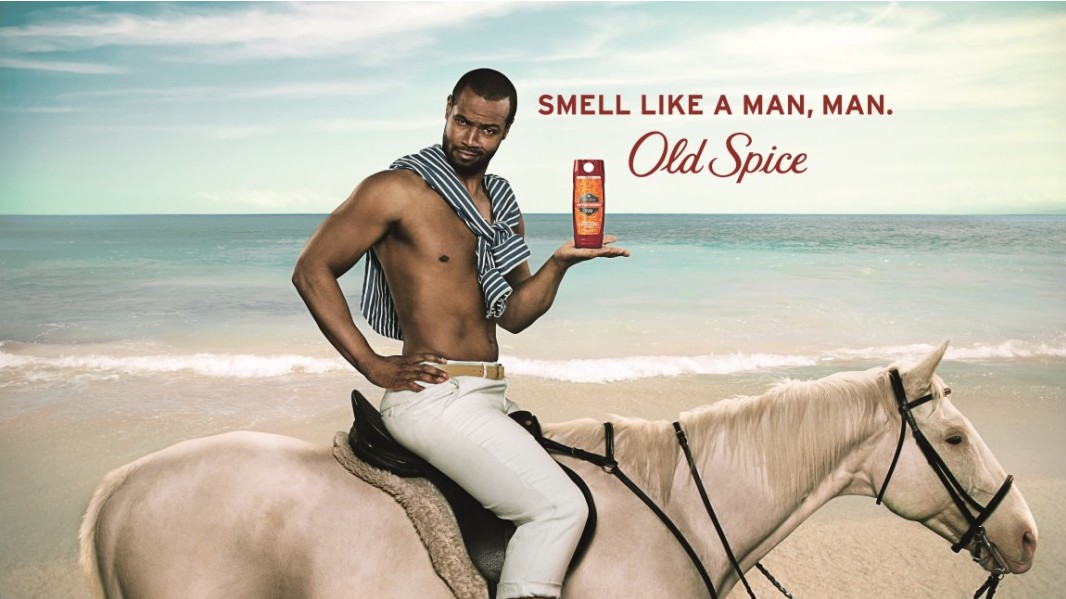}
         \caption{}
         \label{fig:old-spice}
     \end{subfigure}
     \begin{subfigure}[b]{0.12\textwidth}
         \centering
         \includegraphics[width=\textwidth,scale=0.68]{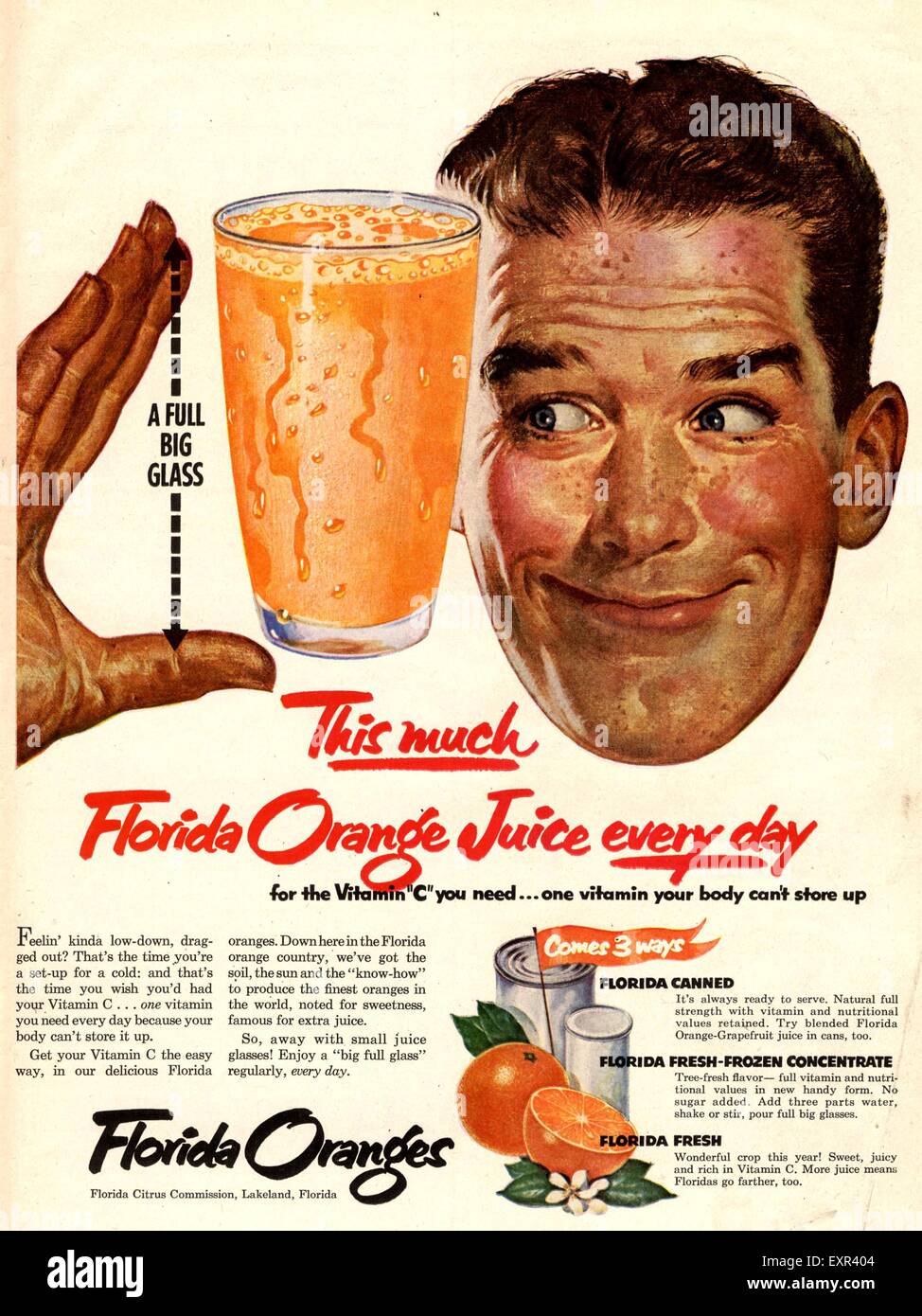}
         \caption{}
         \label{fig:florida-juice}
     \end{subfigure}
     \begin{subfigure}[b]{0.12\textwidth}
         \centering
         \includegraphics[width=\textwidth,scale=0.68]{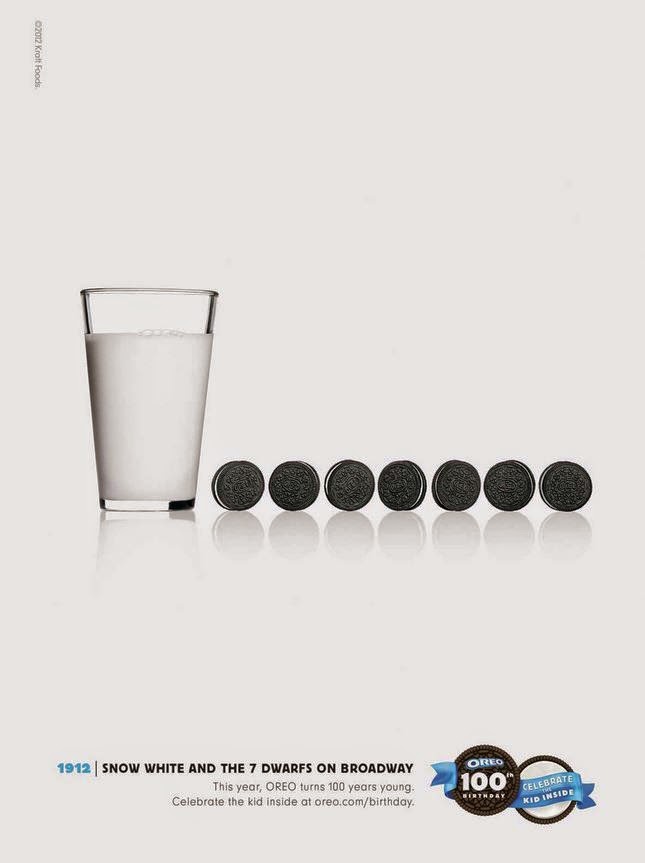}
         \caption{}
         \label{fig:oreo}
     \end{subfigure}

    \caption{Various rhetoric strategies used in advertisements}
    \label{fig:rhetoric-strategies-ads}
\end{figure*}

\begin{figure}
    \centering
    \begin{subfigure}[b]{0.49\textwidth}
         \centering
         \includegraphics[width=0.8\textwidth,scale=0.68]{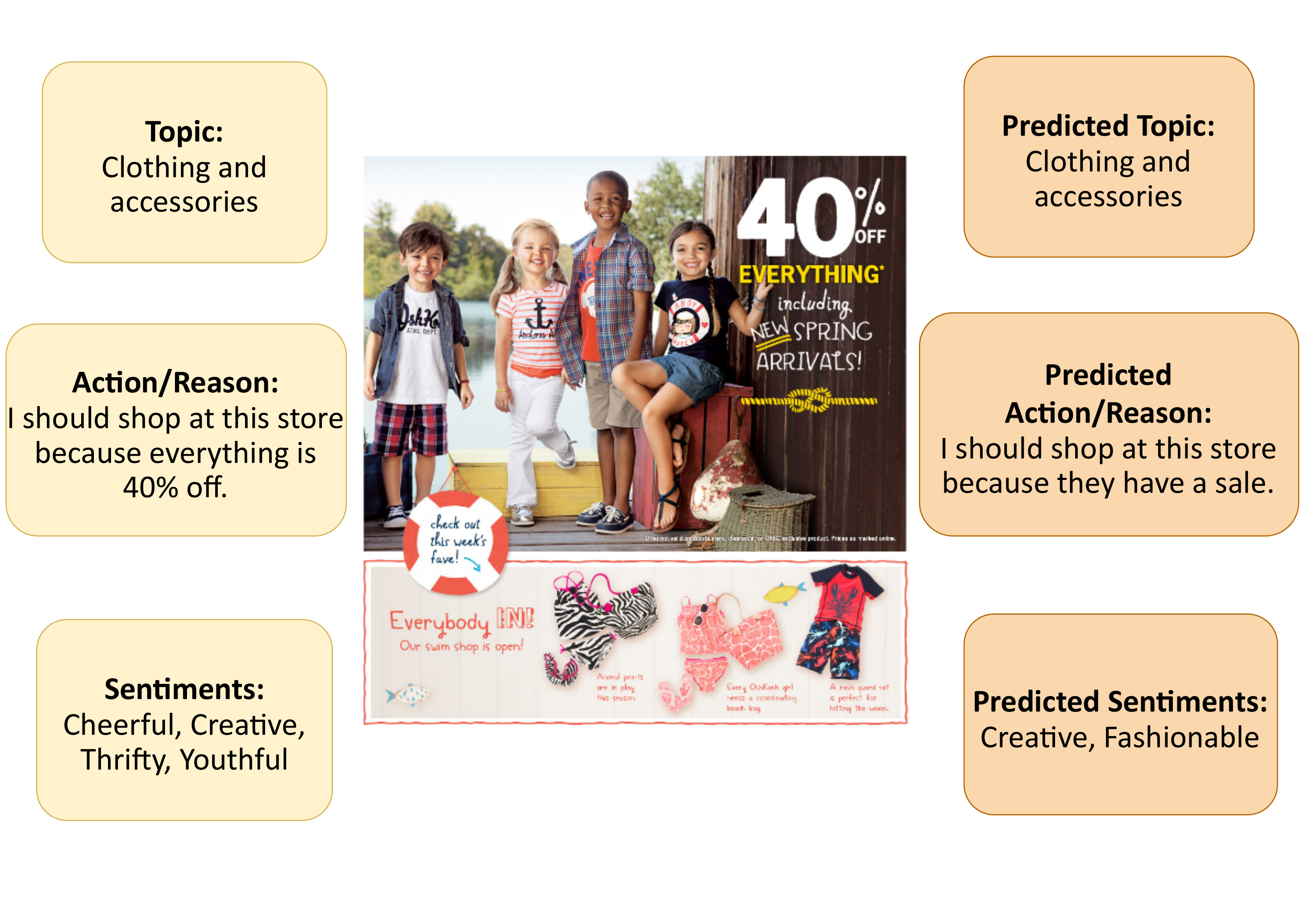}
         \caption{}
         \label{fig:preds-sup-1}
     \end{subfigure}
     \begin{subfigure}[b]{0.49\textwidth}
         \centering
         \includegraphics[width=0.8\textwidth,scale=0.68]{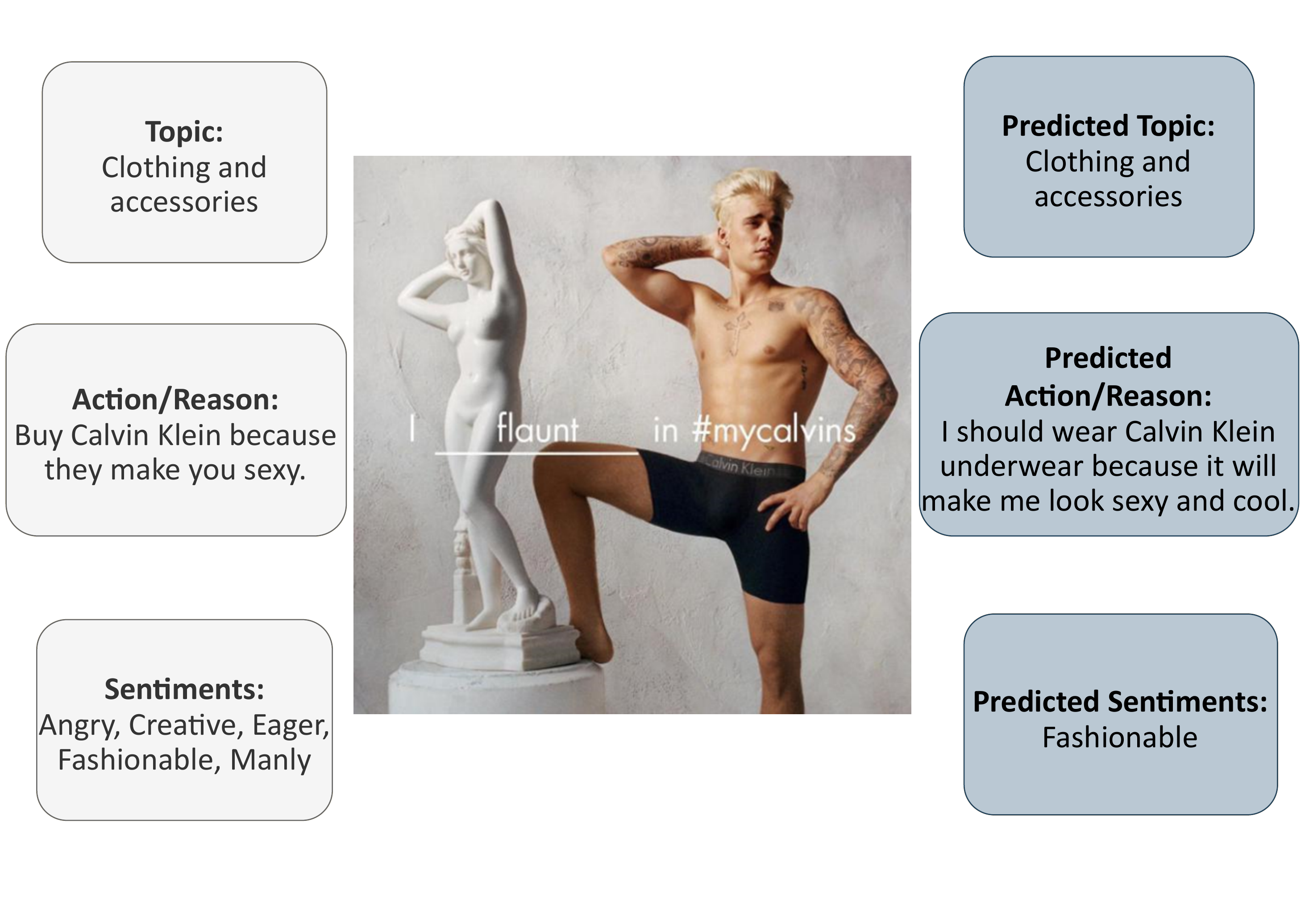}
         \caption{}
         \label{fig:preds-sup-2}
     \end{subfigure}
     \begin{subfigure}[b]{0.49\textwidth}
         \centering
         \includegraphics[width=0.8\textwidth,scale=0.68]{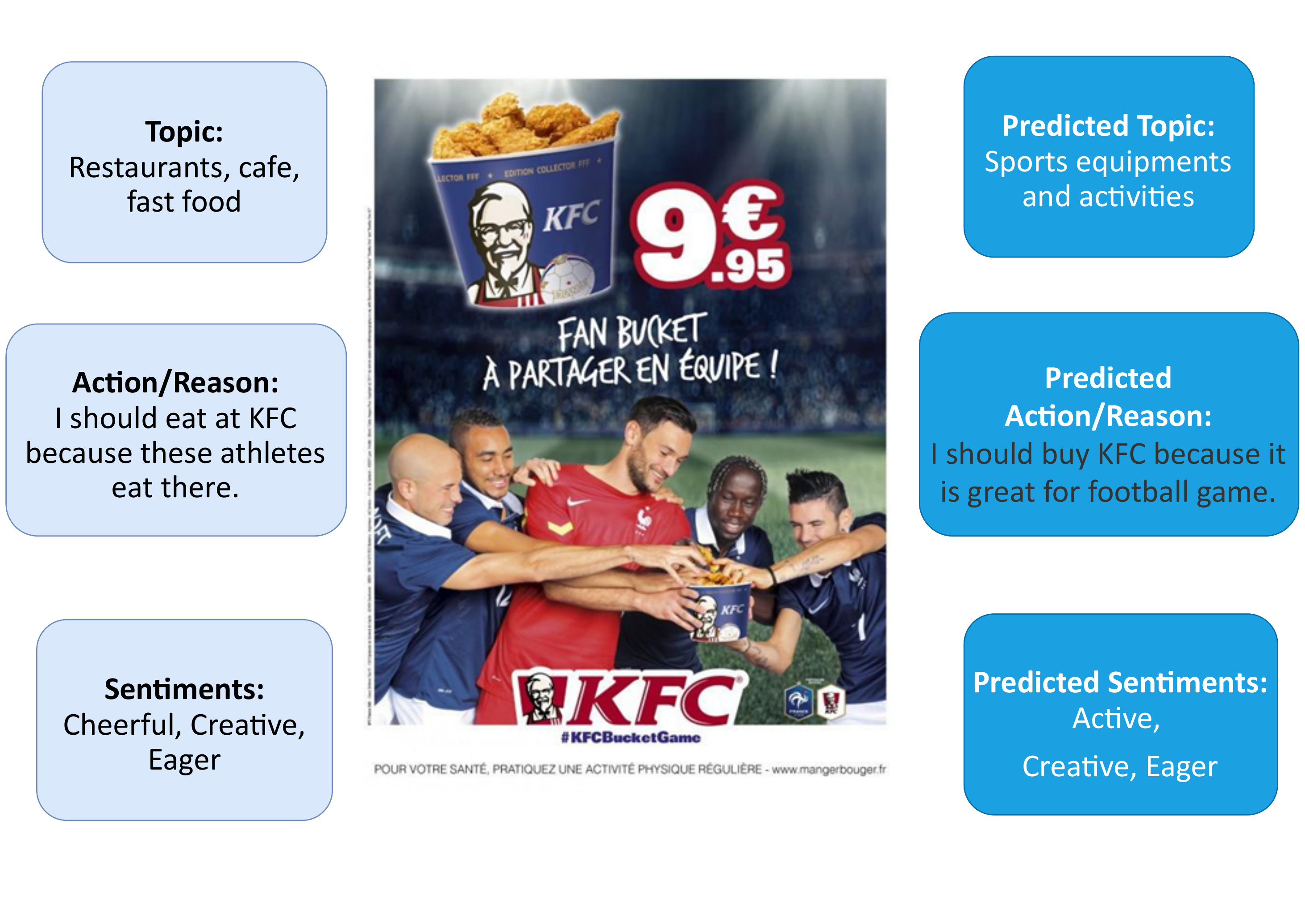}
         \caption{}
         \label{fig:preds-sup-8}
     \end{subfigure}
    \caption{Some samples from the Pitts Ads dataset along with the ground truth and predicted action-reason statement, topic and sentiment.}
    \label{fig:preds-sup}
\end{figure}

\begin{figure}
    \centering
    \begin{subfigure}[b]{0.22\textwidth}
         \centering
         \includegraphics[width=\textwidth,scale=0.6]{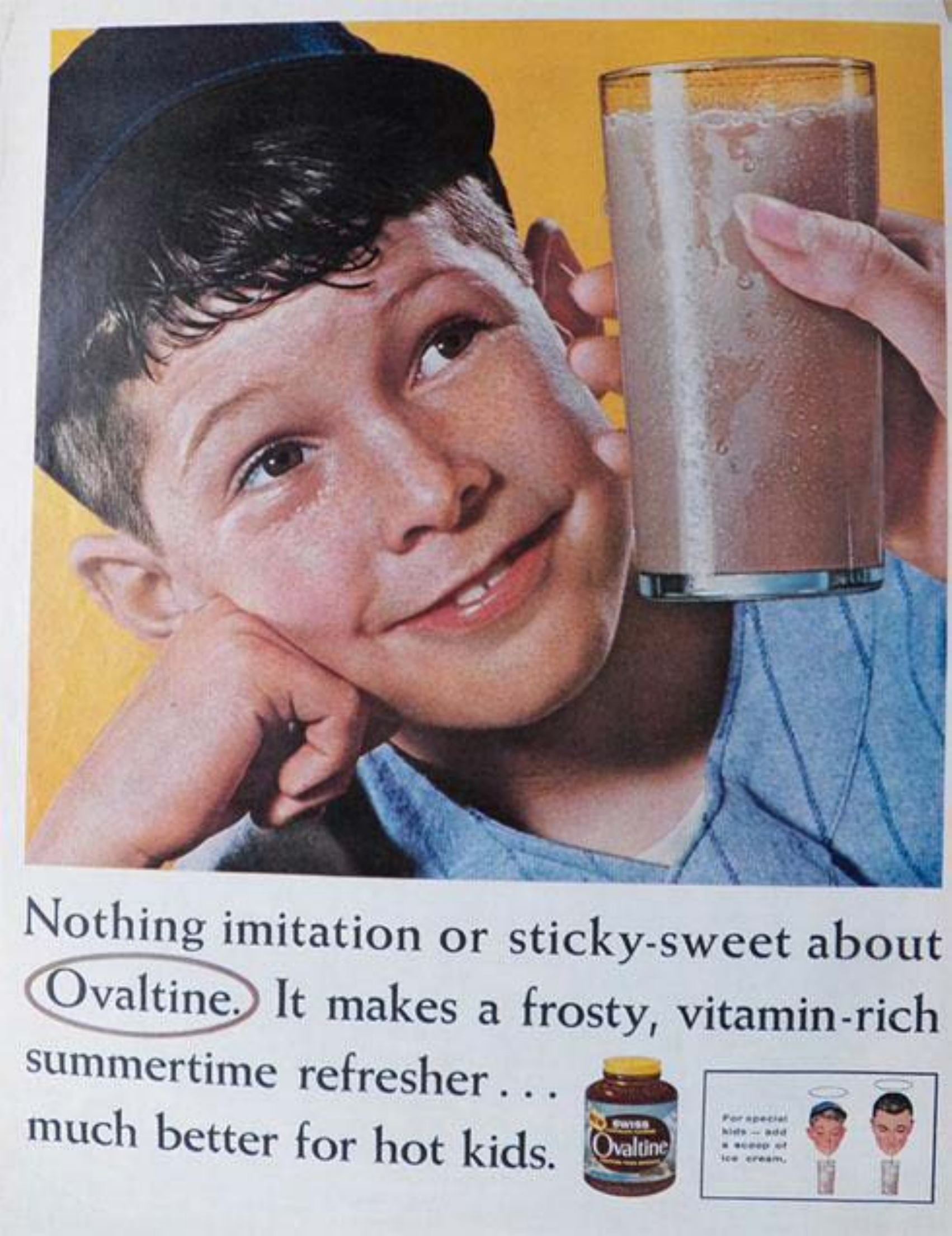}
         \caption{}
     \end{subfigure}
     \begin{subfigure}[b]{0.22\textwidth}
         \centering
         \includegraphics[width=\textwidth,scale=0.5]{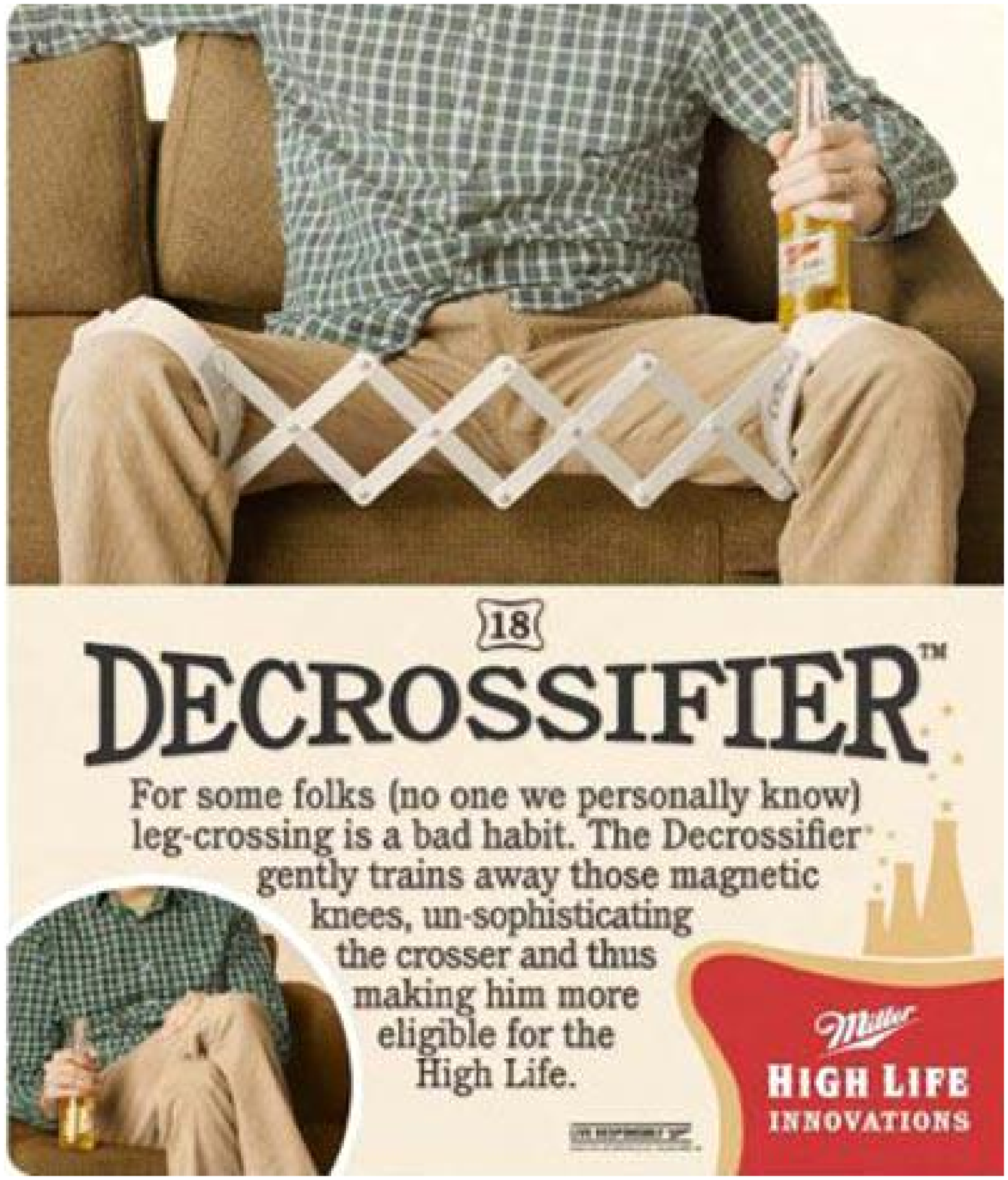}
         \caption{}
         
     \end{subfigure}
     \begin{subfigure}[b]{0.2\textwidth}
         \centering
         \includegraphics[width=\textwidth,scale=1.0]{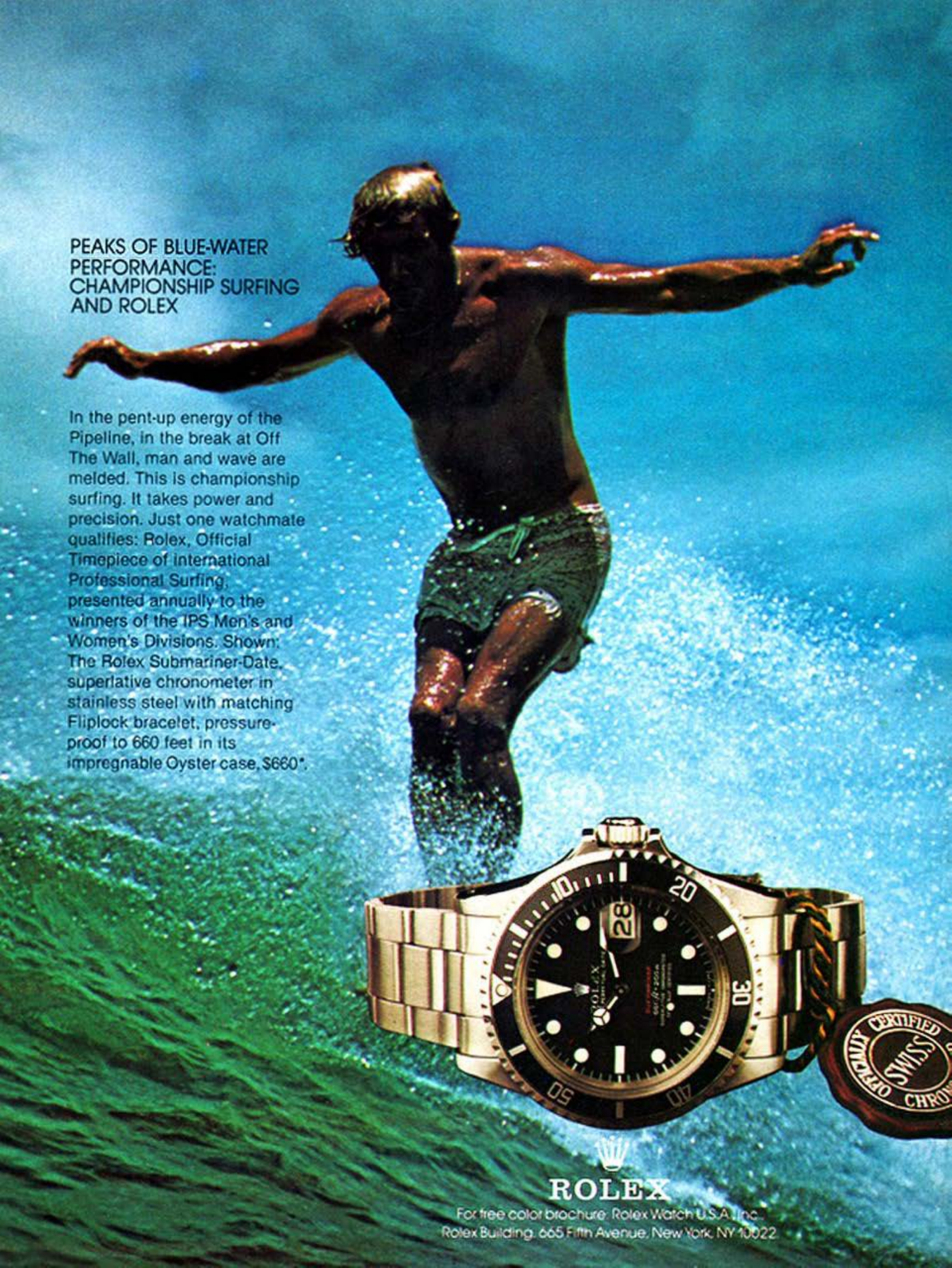}
         \caption{}
     \end{subfigure}
     \begin{subfigure}[b]{0.28\textwidth}
         \centering
         \includegraphics[width=\textwidth,scale=1]{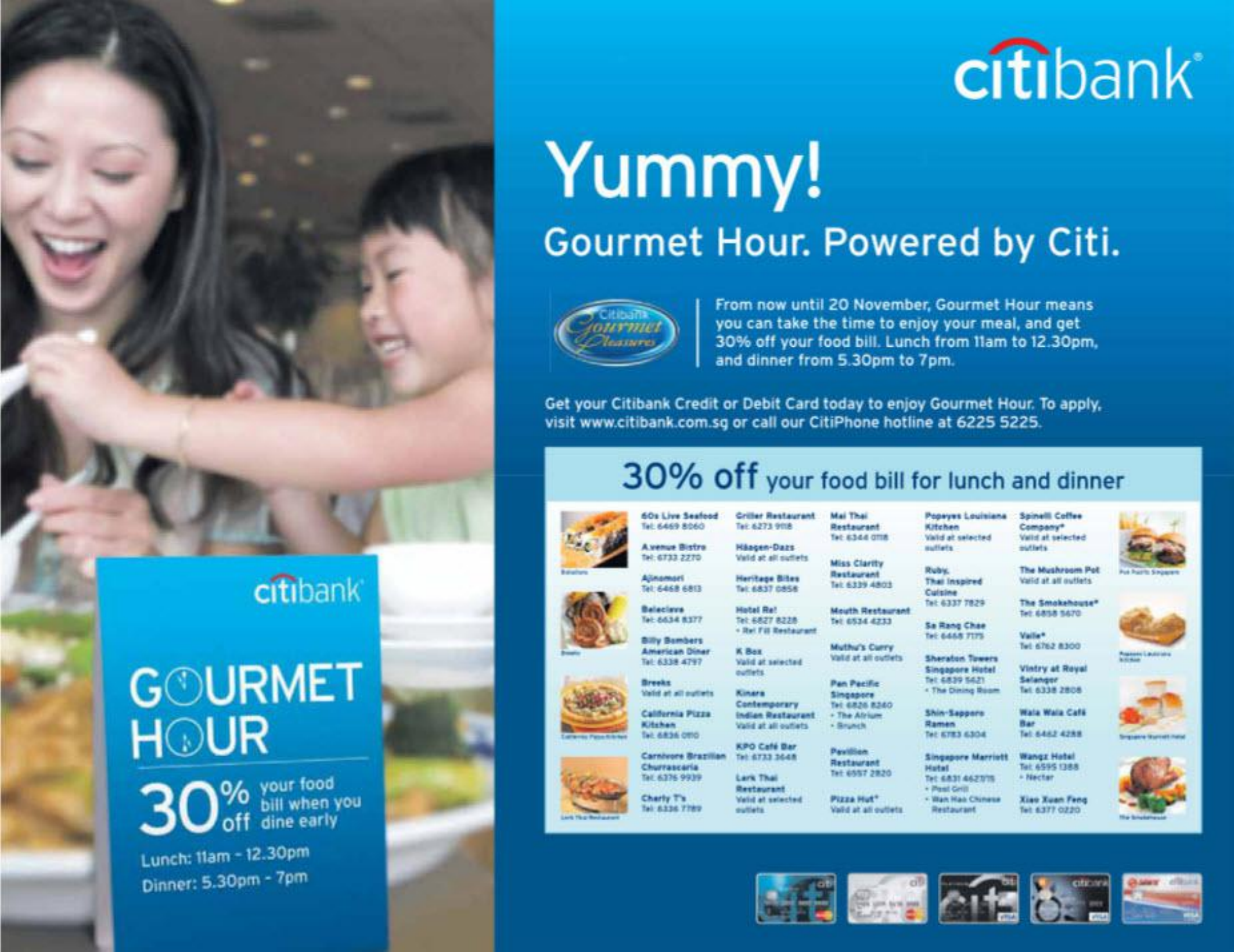}
         \caption{}
         
     \end{subfigure}
    \caption{Advertisements containing humans and concreteness}
    \label{fig:humans_concreteness-ads}
\end{figure}

\begin{figure*}[t]
    \centering
    \begin{subfigure}[b]{0.4\textwidth}
        \includegraphics[scale=0.2]{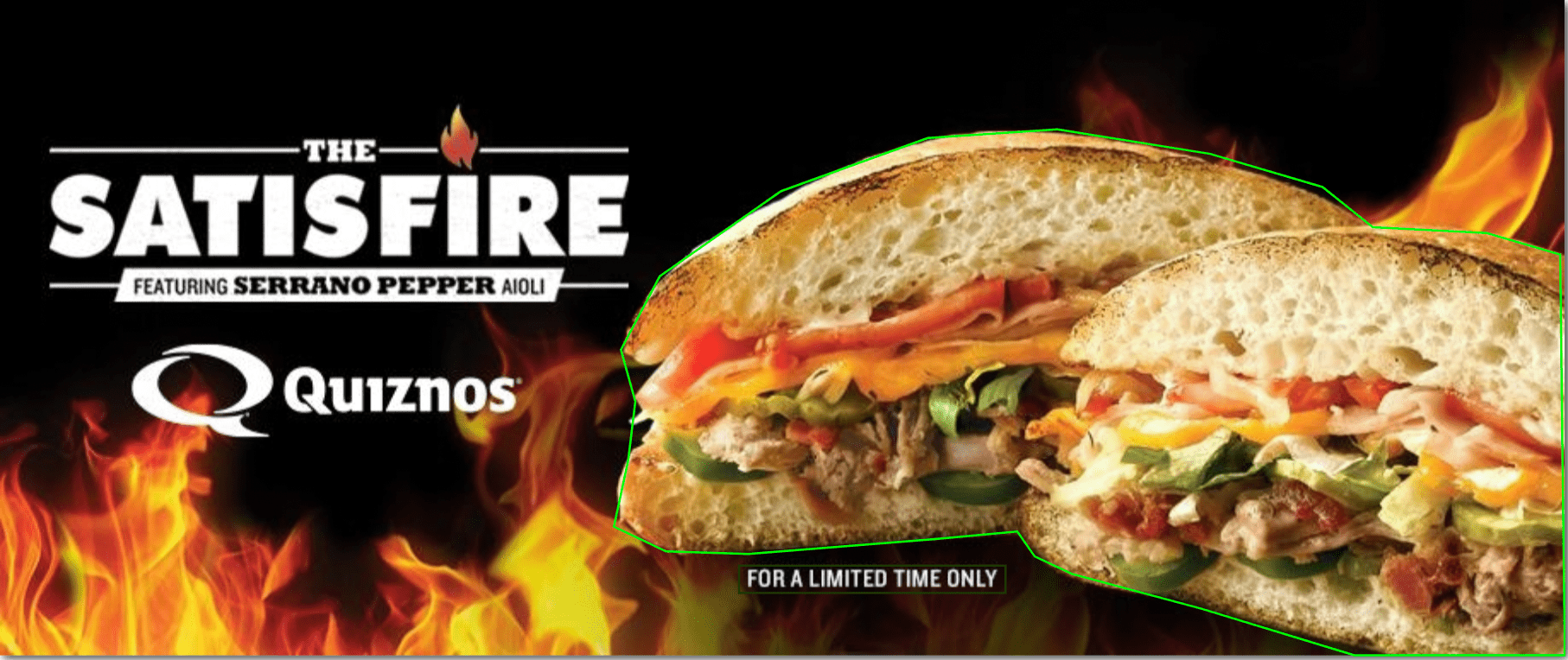}
    \end{subfigure}
    \begin{subfigure}[b]{0.4\textwidth}
        \includegraphics[scale=0.2]{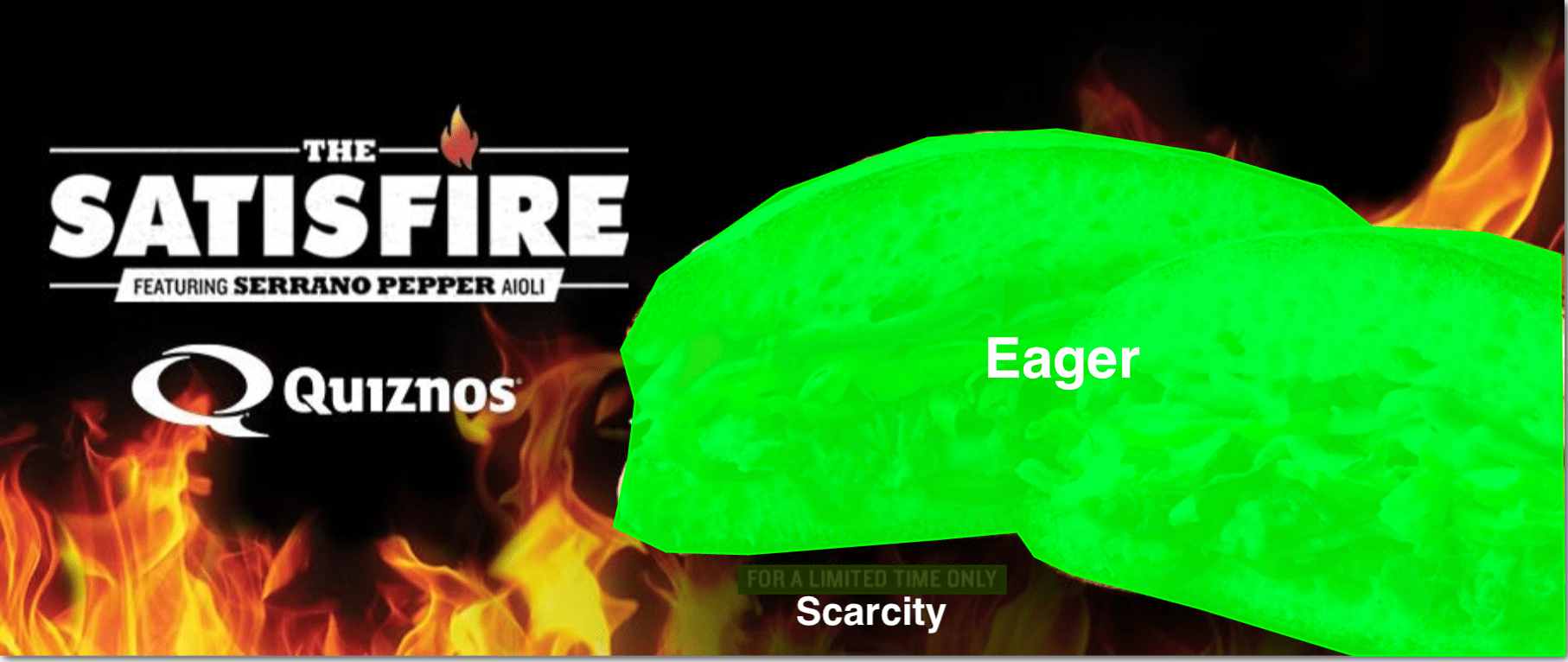}
        
    \end{subfigure}
     \begin{subfigure}[b]{0.45\columnwidth}
         \includegraphics[scale=0.22]{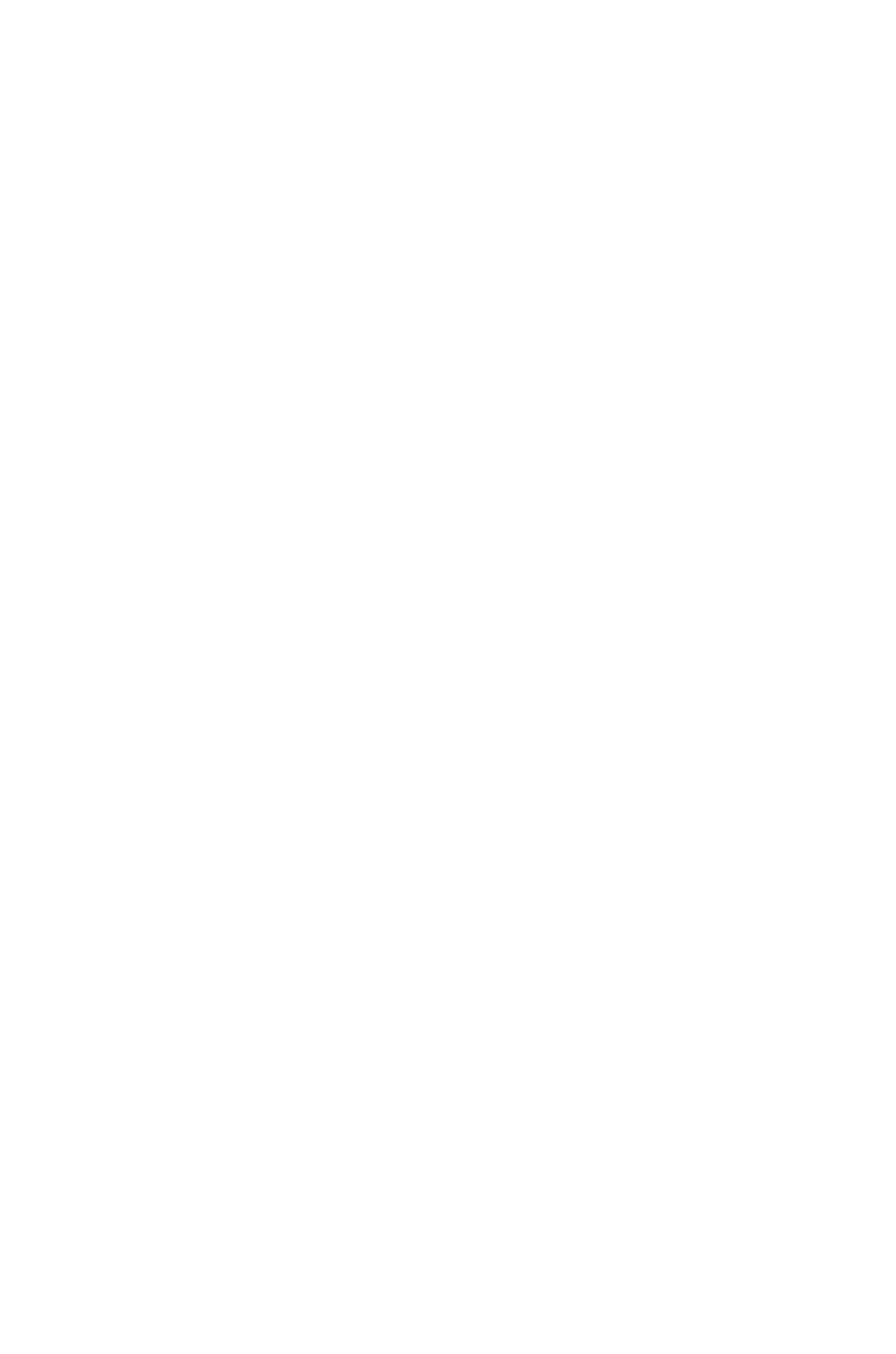}
     \end{subfigure}
     \begin{subfigure}[b]{0.45\columnwidth}
         \includegraphics[scale=0.22]{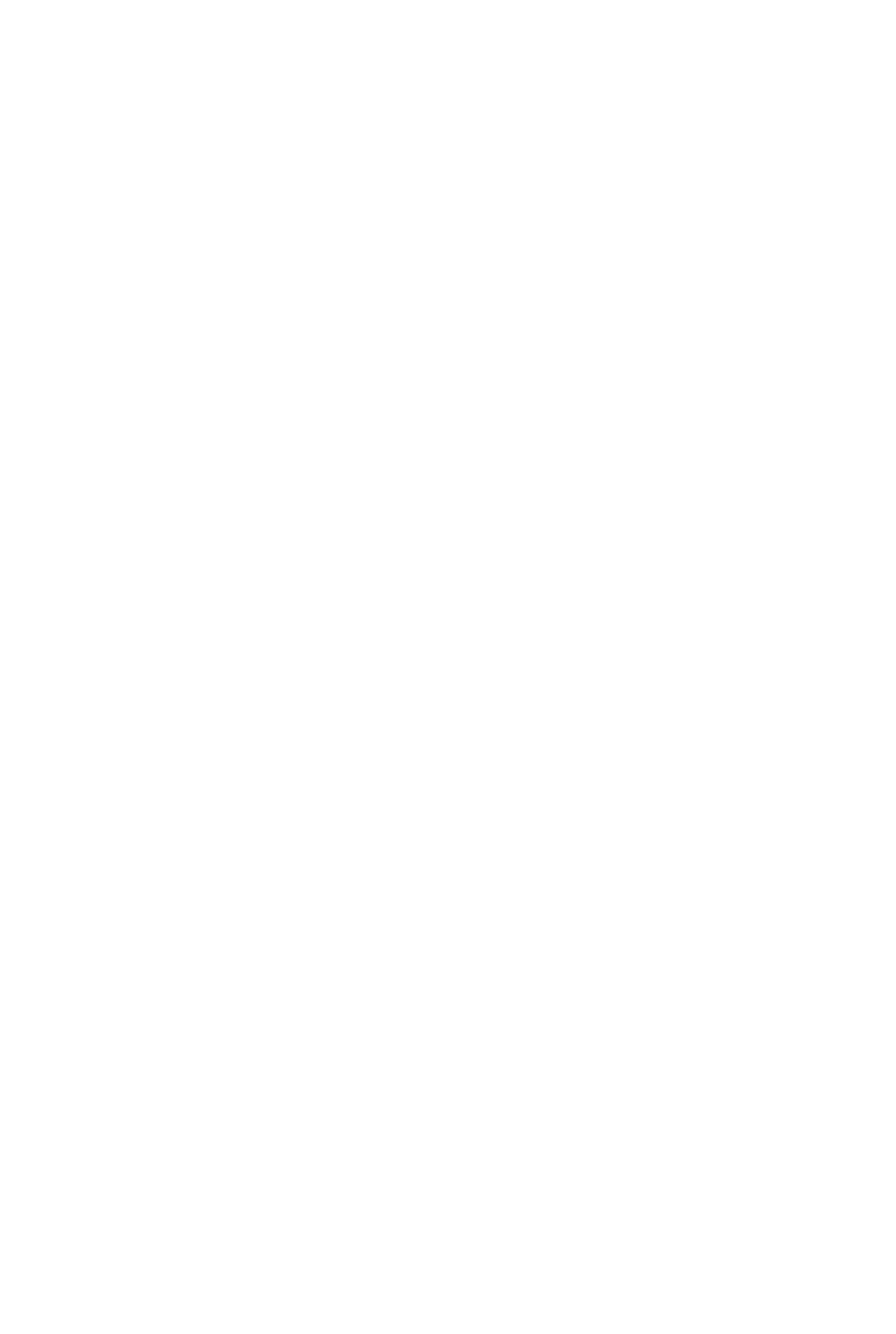}
     \end{subfigure}
     \begin{subfigure}[b]{0.245\textwidth}
         \includegraphics[scale=0.22]{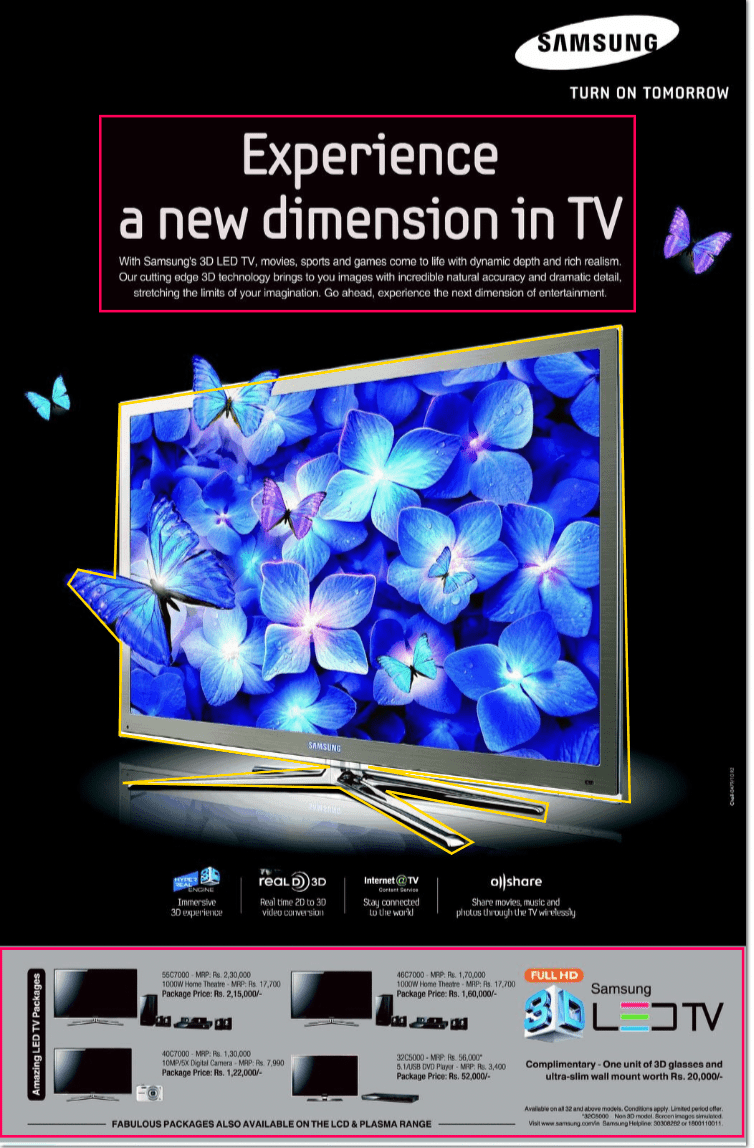}
     \end{subfigure}
     \begin{subfigure}[b]{0.245\textwidth}
         \includegraphics[scale=0.22]{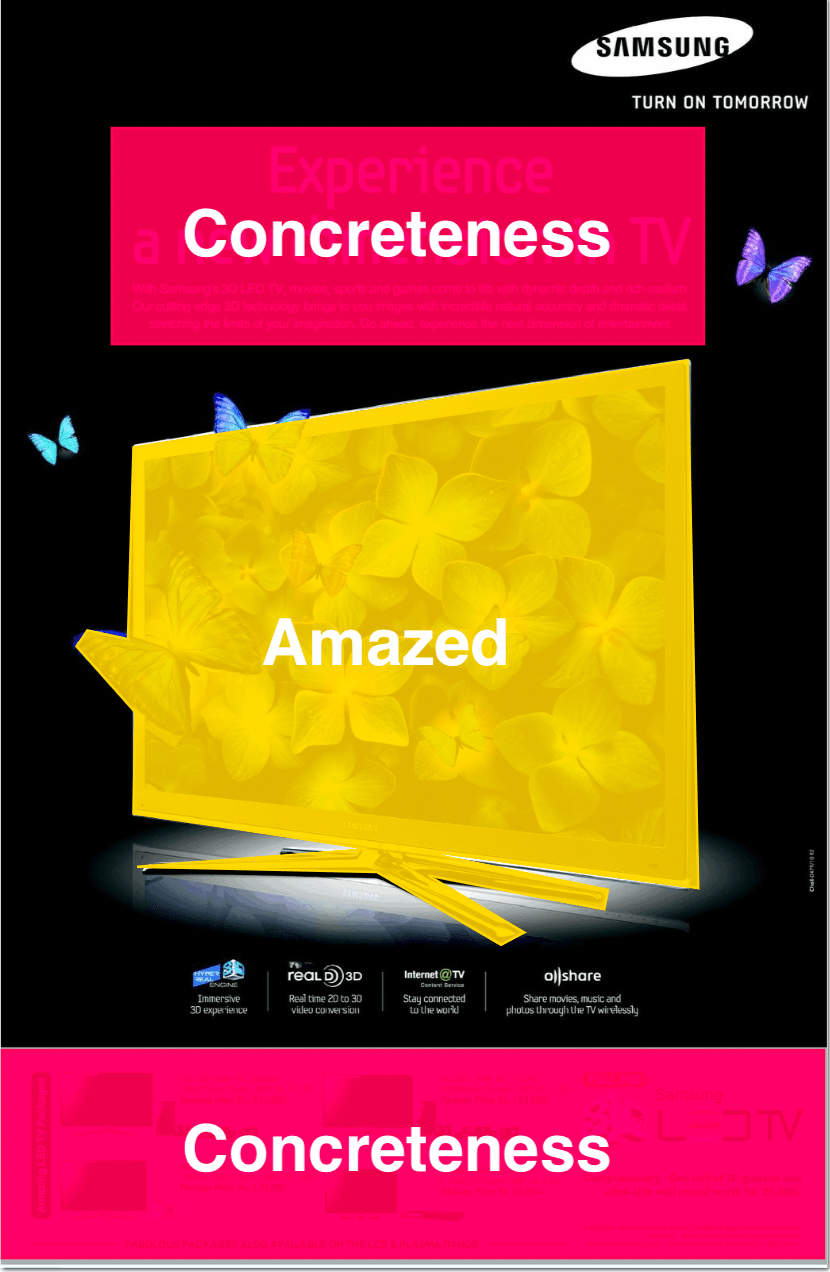}
     \end{subfigure}

     \begin{subfigure}[b]{0.48\columnwidth}
         \includegraphics[scale=0.22]{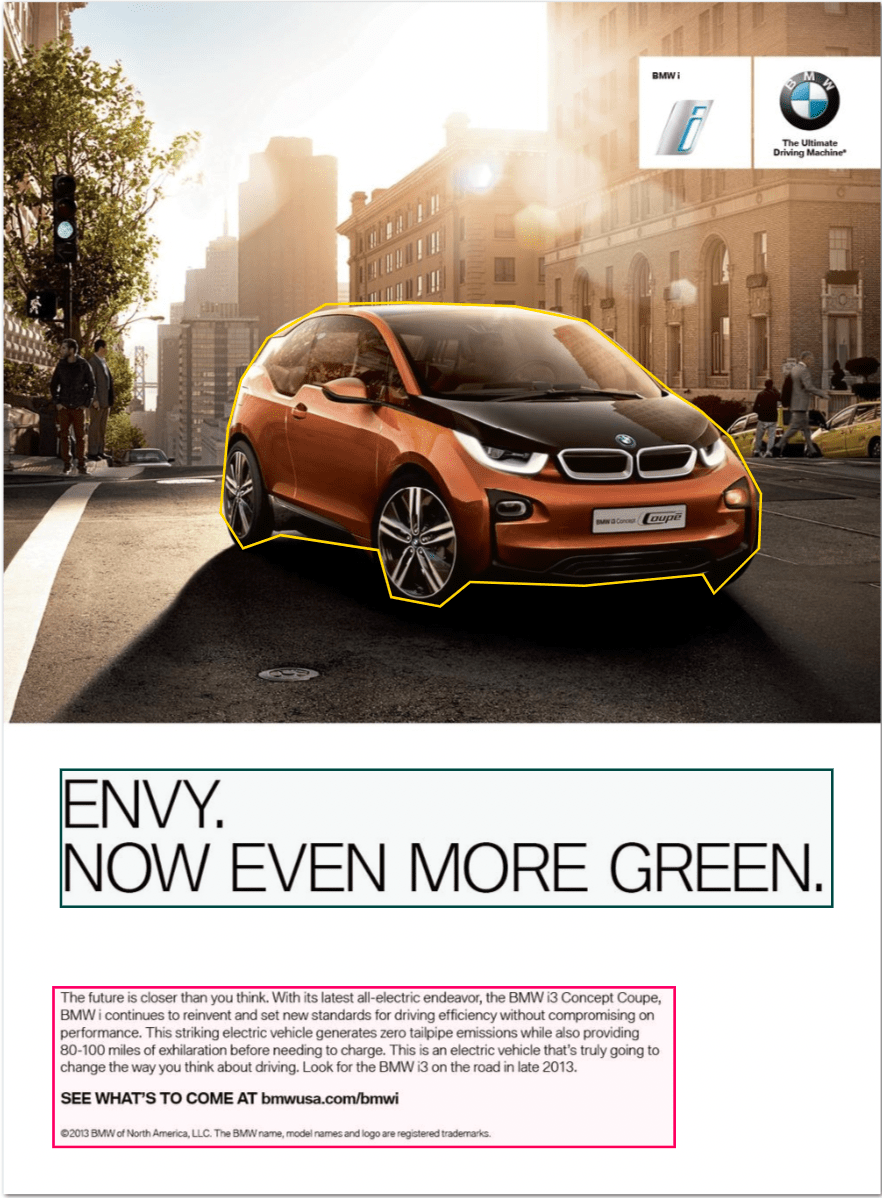}
     \end{subfigure}
     \begin{subfigure}[b]{0.45\columnwidth}
         \includegraphics[scale=0.22]{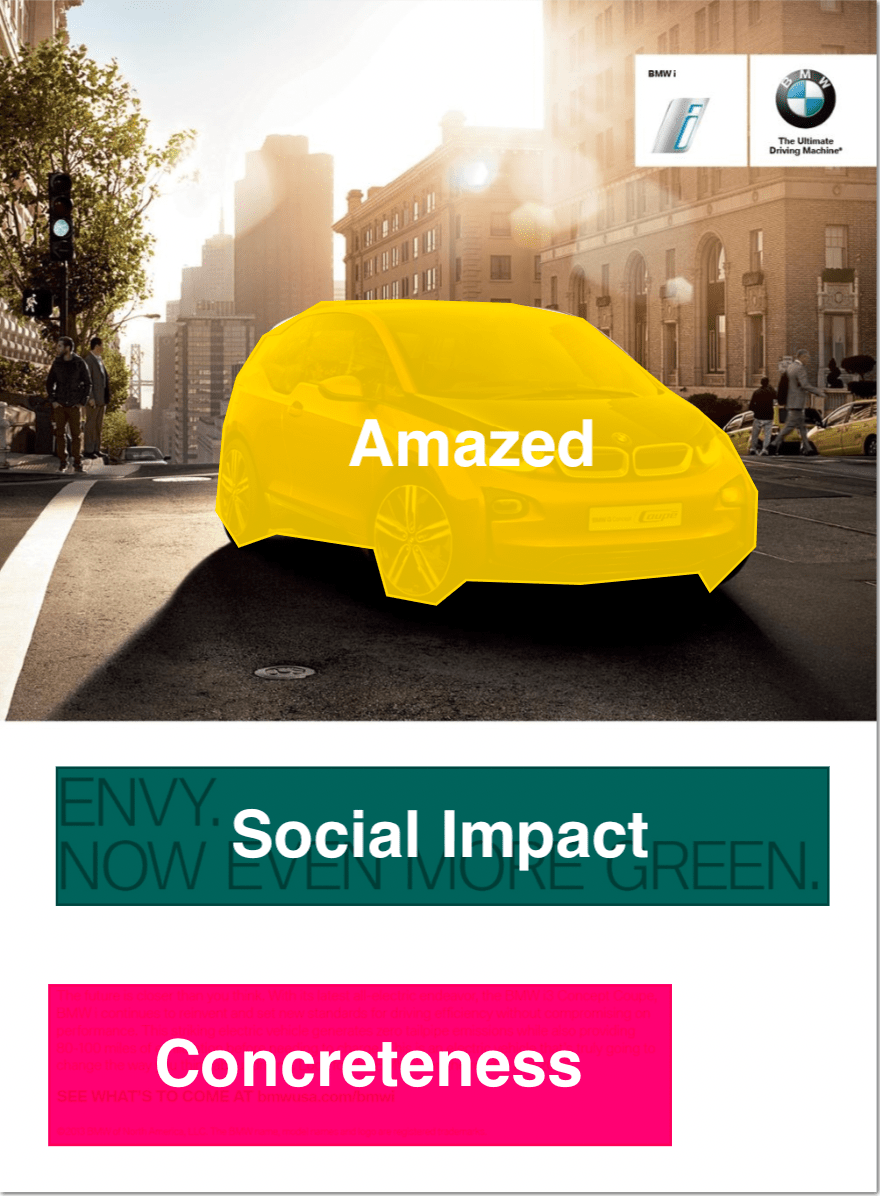}
     \end{subfigure}

     \begin{subfigure}[b]{0.4\textwidth}
         \includegraphics[scale=0.22]{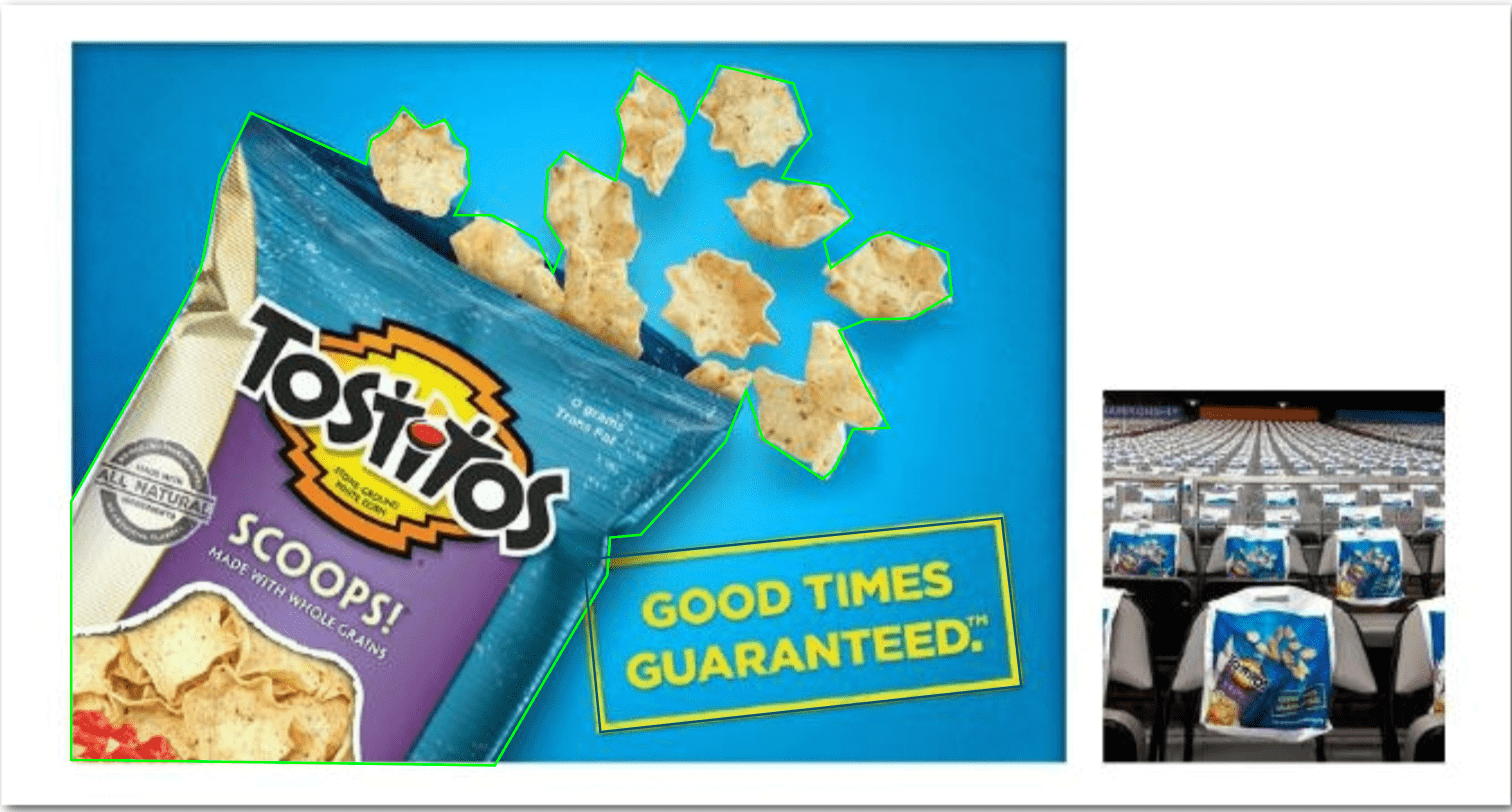}
     \end{subfigure}
     \begin{subfigure}[b]{0.4\textwidth}
         \includegraphics[scale=0.22]{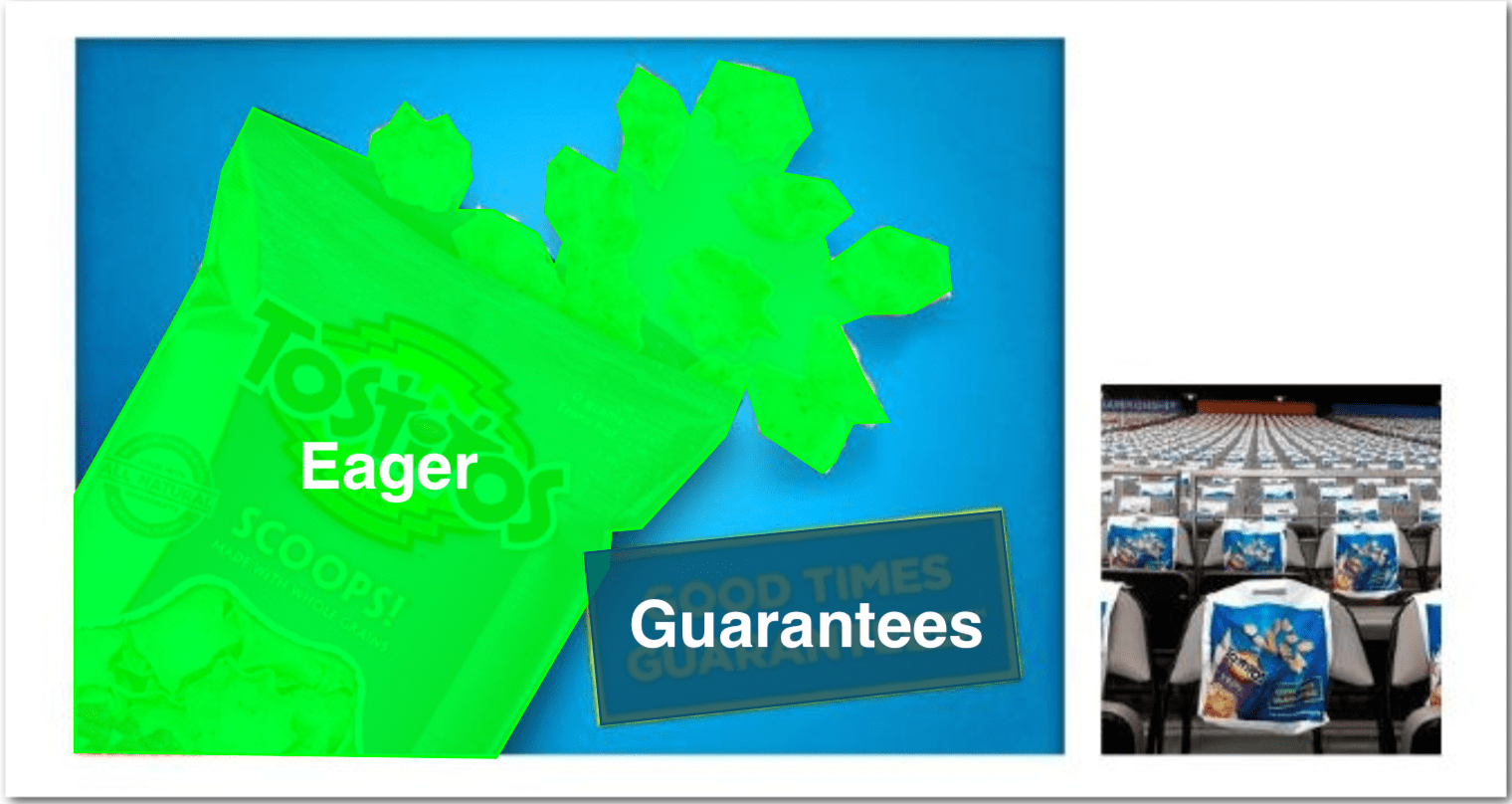}
     \end{subfigure}
     
    \caption{\small \label{fig:appendix segmented image} Image with a segmentation mask depicting the strategies \textit{Emotion:Cheerful}, \textit{Emotion:Eager} and \textit{Trustworthiness}.}
\end{figure*}

\begin{figure*}[t]
        \centering
        \includegraphics[scale=0.6]{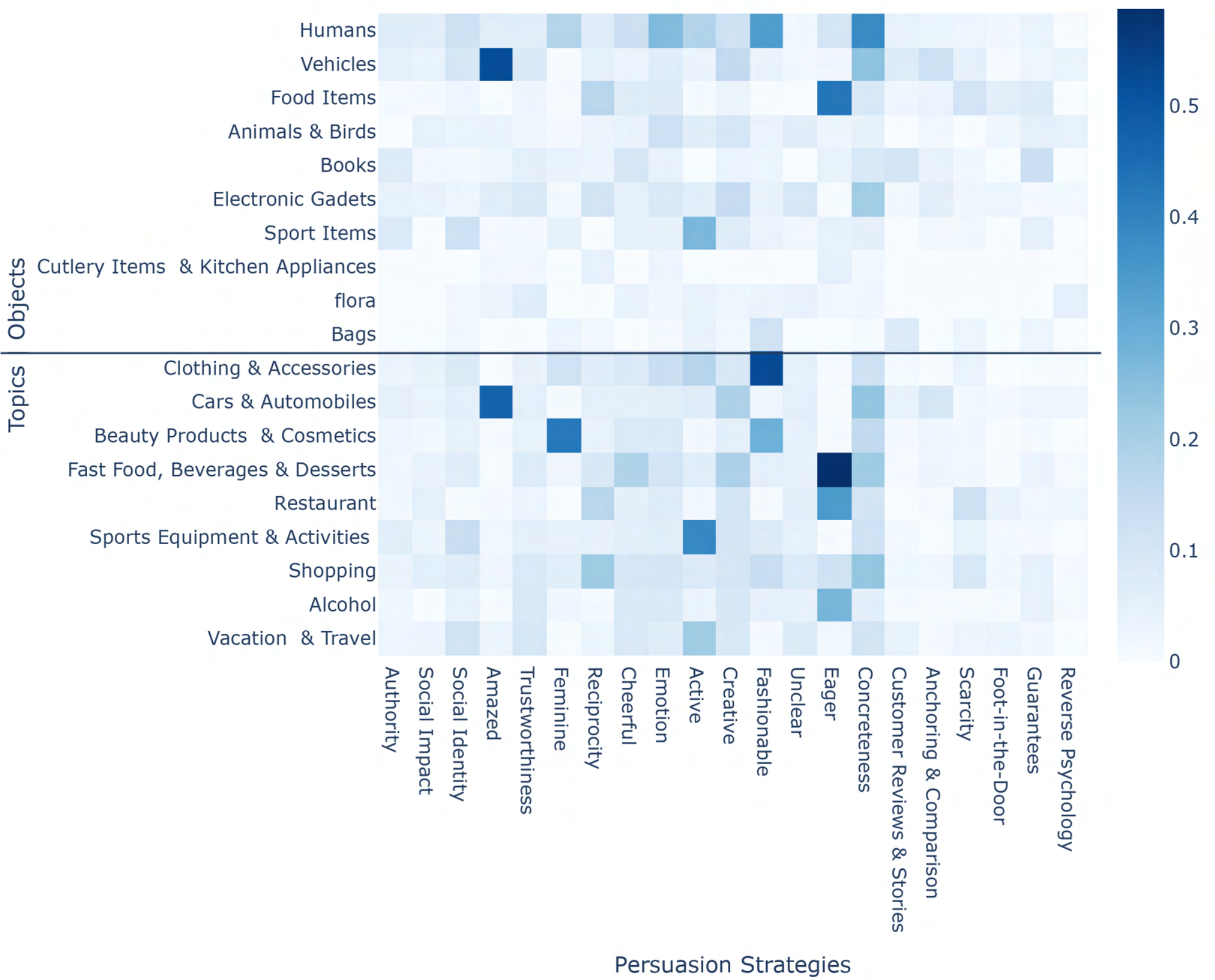}
        \caption{Dice correlation between topics and strategies. Topics are taken from the Pitts Ad dataset and further similar topics are combined to get these values.}
        \label{fig:correlation-between-topics-and-strategies}
    \end{figure*}

\begin{figure*}[t]
        \centering
        \includegraphics[scale=0.8]{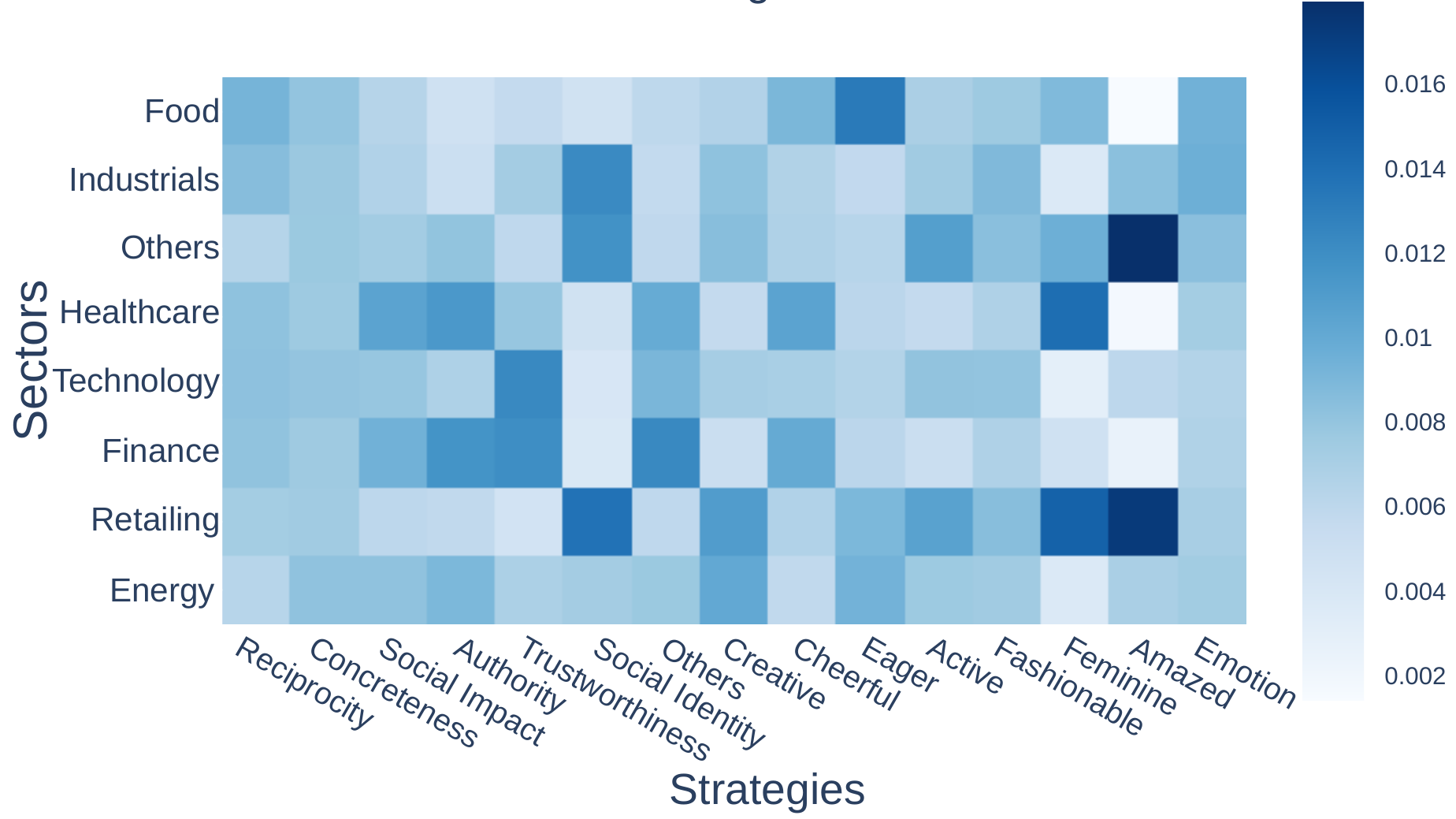}
        \caption{Dice correlation between company sectors and strategies. Companies are taken from the Meta advertisement data and further these companies are labeled according to which sector they belong.}
        \label{fig:correlation-between-company-sector-and-strategies}
    \end{figure*}

\end{document}